\newcommand{\whichFig}{3} 

\newcommand{\whichHyp}{1} 

\documentclass[letterpaper, 10 pt, journal, twoside]{IEEEtran} 

\usepackage{amsmath,amsfonts}
\usepackage{algorithmic}
\usepackage{algorithm}
\usepackage{array}
\usepackage[caption=false,font=normalsize,labelfont=sf,textfont=sf]{subfig}
\usepackage{textcomp}
\usepackage{stfloats}
\usepackage{url}
\usepackage{verbatim}
\usepackage{graphicx}
\usepackage{cite}

\if\whichFig1
    \usepackage{epstopdf}
    \AppendGraphicsExtensions{.tif}
\fi

\usepackage{ctable} 

\begin{document}

\title{How Does the Inner Geometry of Soft Actuators Modulate the Dynamic and Hysteretic Response?}


\author{Jacqueline Libby, \textit{Member, IEEE}$^{1}$, Aniket A. Somwanshi$^{2}$, Federico Stancati$^{2}$, Gayatri Tyagi$^{3}$, 
\\ Sarmad Mehrdad, \textit{Student Member, IEEE}$^{2}$, JohnRoss Rizzo$^{1,4}$, S. Farokh Atashzar, \textit{Senior Member, IEEE}$^{1,2,5,6}$
\thanks{\hspace{-2.2em}$^{1}$Center for Urban Science and Progress, Tandon School of Engineering, New York University (NYU), Brooklyn, NY 11201 USA.
        {\tt\small jlibby@nyu.edu}}%
\thanks{\hspace{-2.2em}$^{2}$Dept. of Mechanical and Aerospace Engineering, NYU Tandon}
\thanks{\hspace{-2.2em}$^{4}$Dept. of Computer Science and Engineering, NYU Tandon}
\thanks{\hspace{-2.2em}$^{5}$School of Medicine, NYU, New York, NY, 10016}
\thanks{\hspace{-2.2em}$^{6}$Dept. of Electrical and Computer Engineering, NYU Tandon}
\thanks{\hspace{-2.2em}$^{7}$\tt\small Corresponding author: f.atashzar@nyu.edu, Phone: 646.997.3541, Address: 370 Jay Street, Room 904, Brooklyn, NY 11201}%
}



\maketitle

\begin{abstract}
This paper investigates the influence of the internal geometrical structure of soft pneu-nets on the dynamic response and hysteresis of the actuators.
The research findings indicate that by strategically manipulating the stress distribution within soft robots, it is possible to enhance the dynamic response while reducing hysteresis.
The study utilizes the Finite Element Method (FEM) and includes experimental validation through markerless motion tracking of the soft robot.
In particular, the study examines actuator bending angles up to 500\% strain while achieving 95\% accuracy in predicting the bending angle.
The results demonstrate that the particular design with the minimum air chamber width in the center significantly improves both high- and low-frequency hysteresis behavior by 21.5\% while also enhancing dynamic response by 60\% to 112\% across various frequencies and peak-to-peak pressures.
Consequently, the paper evaluates the effectiveness of "mechanically programming" stress distribution and distributed energy storage within soft robots to maximize their dynamic performance, offering direct benefits for control.
\end{abstract}


\section{Introduction}\label{sec.introduction}

Soft robotics has gained significant attention over the past two decades due to their compliance, safety, low cost, and low weight~\cite{DellaSantina2020-EncyclopediaOfRobotics,Laschi2012-AR}. 
High compliance allows conformity even with delicate materials, which makes them safe for physical human-robot interaction (pHRI) in assistive~\cite{Atashzar2021-FrontRobotAI}, rehabilitative~\cite{Zhao2023-TMECH}, diagnostic~\cite{Coevoet2022-RAL}, and surgical~\cite{Ranzani2016-TRO} applications.
Their safety and affordability will ultimately enhance accessibility to such technologies~\cite{Polygerinos2015-RobAutonSyst}.
Soft Fluidic Actuators (SFAs) are the most prevalent building blocks of soft robots, powered pneumatically or hydraulically and made from deformable inexpensive elastomeric materials~\cite{Rus2015-Nat}.
Pneu-nets are a family of SFAs composed of networks of air chambers%
~\cite{Ilievski2011-AngewChemIntEd}.
Pneu-nets are easy to fabricate and are inflatable with low pressures.

Closed-loop control of soft robots (specifically pneumatic soft robots) can be challenging due to (a) the strong nonlinearities and complex dynamics arising from the hyperelasticity of soft materials, (b) the compressibility of air in pneumatic actuation, (c) the lag introduced from valve commands, and (d) the difficulty in sensing complex continuum shape configurations~\cite{Skorina2015-ICRA,Chen2020-RAL,Li2022-NatRevMater,Xavier2022-TMECH}.
In some applications, such as soft grippers (used for handling delicate objects with the binary task of grip or release), only simple control is necessary to achieve compliant under-actuated grasping of any shape.
However, for pHRI, control is not as simple since safety is a more important requirement, and system tasks are more complicated.
This paper focuses on understanding the interplay between the internal geometry of soft robots and important control features, including maximum dynamic response, repeatability, and hysteresis. The outcome sheds light on how to design robots with high dynamic response while minimizing hysteresis to reduce the complexity of control.

There has been research in characterizing the dynamics of pneu-nets to introduce new sensing schemes~\cite{Park2019-RoboSoft,Giffney2016-Robotics}
and new control algorithms~\cite{Alian2022-TMECH}. 
There has also been work in characterizing dynamics for geometrically-altered designs.
Mosadegh~\emph{et al}.~\cite{Mosadegh2014-AdvFunctMater} measured hysteresis and speed to evaluate the performance of the fast pneu-net (fPN).
Variants of the fPN have been proposed to optimize speed, force, and system response time~\cite{Park2018-RAL,Lotfiani2018-ICIRA}.
Recently we introduced a mechanical coding technique for multi-section 3D printed pneu-nets and discussed the hysteresis of various codings~\cite{Altelbani2021-RAL}.
\IEEEpubidadjcol

Geometrical design optimization of pneu-nets has been an active line of research to enhance static metrics such as bending angle, force, contact area, or friction~\cite{Hu2018-Robotics,Liu2020-TMECH-B,Zhou2019-SaeIntJMaterManuf,Liu2020-TMECH-A,Tennakoon2021-ICAR,Venkatesan2021-WorldJEng,Thompson-Bean2016-BioinspirBiomim}.
Mechanical parameters considered include (a) passive layer thickness, (b) geometry of the channel cross-section, (c) number of chambers, (d) number of sections in multi-section pneu-nets, and (e) interior wall thickness and the gap between chambers in fPNs.
Some parameters are held constant within each design iteration~\cite{Hu2018-Robotics,Liu2020-TMECH-B}, while other parameters are varied along the body of the actuator~\cite{Zhou2019-SaeIntJMaterManuf,Liu2020-TMECH-A,Tennakoon2021-ICAR,Venkatesan2021-WorldJEng,Thompson-Bean2016-BioinspirBiomim}.

Although the above body of work is rich in design optimization, it does not explore how design affects dynamic performance (except for \cite{Liu2020-TMECH-A}, which measures response time) and how this performance corresponds to the propagation of stress in the actuator.
Liao~\emph{et al}.~\cite{Liao2020-PolymTest} report that Ecoflex silicone (the silicone used herein) has minimal hysteresis after a few loading/unloading cycles.
Therefore, the hysteresis observed in actuators made from Ecoflex is an effect of the geometry, not the material.
This fact makes optimizing the geometrical design for hysteretic behavior even more critical.

Shape sensing is required to study dynamic behavior.
For soft actuators, a common first-order metric of shape is the bending angle.
For soft actuators that are hydraulically actuated, with design constraints that only allow bending about one axis ~\cite{Mosadegh2014-AdvFunctMater,Park2018-RAL}, the internal volume is highly correlated with bending angle, so a direct bending angle measurement can be replaced by a volume measurement.
In pneumatic actuation, the compressibility of air leads to more complex volume measurements.
The work in~\cite{Deimel2021-Thesis} derives air mass from sensed pressure to deduce a soft actuator's volume, but the results showed significant drift.
Embedded strain sensing~\cite{Park2019-RoboSoft,Giffney2016-Robotics} relies on more complex fabrication and is limited to soft designs that include a passive layer for curvature measurement.
Temporarily-attached inertial sensors~\cite{Caasenbrood2022}, goniometers~\cite{Altelbani2021-RAL}, and markers rely on secure mounting points, which becomes complicated for softer materials.
Hence, camera imagery is often used.
Many of the works cited above manually extract bending angles from images~\cite{Hu2018-Robotics,Liu2020-TMECH-B,Liu2020-TMECH-A,Thompson-Bean2016-BioinspirBiomim,Lotfiani2018-ICIRA,Liu2020-TMECH-A}, but this is only sufficient when fine temporal resolution is not required, such as for static metrics or simple dynamic metrics like response time. 

A solution to evaluate dynamic behavior of soft actuators using curvature measurements is automated markerless vision-based techniques.
Such techniques, which mostly use deep learning, have recently arisen for estimating human gesture and animal pose~\cite{Mathis2018-NatNeurosci,Pereira2022-NatMethods}. 
Some recent work has been done to assess the feasibility of utilizing such tools to estimate the pose of soft robots.
In one case, such a method was used to assist with marker-based motion tracking~\cite{Niu2021-Soro}, and in another, to track a robotic gecko's tail reflex~\cite{Siddall2021-CommunBiol}.
More investigation into the usability of markerless vision systems is needed for decoding the dynamic behavior of soft robots.

In this paper, for the first time, we optimize the mechanical design of slow pneu-nets to minimize hysteresis and maximize dynamicity.
For this, we introduce a novel family of designs, allowing a set of parameters to vary linearly along the longitudinal axis.
However, instead of monotonic variation as in~\cite{Zhou2019-SaeIntJMaterManuf,Liu2020-TMECH-A,Tennakoon2021-ICAR,Venkatesan2021-WorldJEng}, we opt for a non-monotonic variation and alter the parameters towards the center and then reverse the direction of variation towards the other end.
In this way, we take the next step of changing the function of parameter variation from linear to piecewise linear and symmetric, drawing attention to this function as an intentional design choice.
\if\whichHyp0
In this paper, nine structural variations of this design family are studied to evaluate the effect of different mechanical codings on the metrics of stress distribution, dynamic response, and hysteresis.
Consequently, correlations between these metrics are observed.
\fi
\if\whichHyp1
Our hypothesis is that decentralizing the stress along the actuator body would decentralize the stored potential elastic energy during actuation, which in turn could alter energy dissipation and thus alter hysteresis.
Consequently, in this paper, nine structural variations of this design family are studied to evaluate the effect of different mechanical codings on redistributing stress and altering the dynamic behavior of the actuators.
\fi

In this paper, to automatically quantify bending angle with fine temporal resolution, we utilize the DeepLabCut (DLC)~\cite{Mathis2018-NatNeurosci} tool, a computer vision algorithm that allows for markerless tracking of large curvatures.
Using the power of deep learning in DLC and data collected in this work (which will be released at the time of publication), this work implements a markerless vision-based assessment platform that evaluates the hysteresis behavior and dynamic response of soft robots.
The results show that the design whose air chamber width decreases towards the center (referred to as \emph{mid-width smaller} in Fig.~\ref{fig.fabNine}) has the best dynamic performance in terms of reduction in hysteretic behavior and increase in dynamic response (maximum bending).

Our FEM analysis takes advantage of a 3rd-order nonlinear polynomial model for material property and successfully simulates the internal stress of soft robots subject to pressurized air with over 500\% deformation.
\if\whichHyp0
The computational analysis powered by FEM showed that the (the \emph{mid-width smaller} design) had the best dynamic performance in terms of dynamic response and hysteretic behavior.
Additionally, this design had the lowest stress in the center of the actuator, suggesting a correlation between minimized central stress and dynamic performance.
\fi
\if\whichHyp1
The computational analysis powered by FEM showed that the (the \emph{mid-width smaller} design) had the best dynamic performance in terms of dynamic response and hysteretic behavior.
Additionally, this design had the lowest stress in the center of the actuator, suggesting a correlation between minimized central stress and dynamic performance.
\fi

The three contributions of this paper are summarized below:
\begin{enumerate}
	\item Introduction of a new family of designs with non-monotonic, symmetric variation of parameters to optimize hysteresis of soft actuators while studying dynamicity. 
	\item FEM analysis decoding decentralized distribution of stress, which supports improved dynamic behavior.
	\item Markerless vision-based motion tracking for hysteretic analysis of soft robots.
\end{enumerate}

The rest of this paper is organized as follows.
Sec.~\ref{sec.methods} will discuss the structure of the nine designs, the fabrication pipeline, the construction of the FEM models, the pneumatic control for actuation, and the markerless assessment platform.
Sec.~\ref{sec.results} will present an analysis of the FEM models in terms of bending angle and distributed stress, the static analysis of experimental data to validate the FEM models, and the dynamic analysis of experimental data in terms of dynamic response and hysteretic behavior.
Sec.~\ref{sec.discussion} will discuss the correlation between the static and dynamic analysis, supporting the intuition behind our design optimization.
Sec.~\ref{sec.conclusion} will summarize and contextualize these results.

\section{Methods}\label{sec.methods}

\subsection{Design}\label{subsec.design}  

\begin{figure}[htb]\vspace{-5pt}
    \if\whichFig0
        \includegraphics[width=\linewidth]{figures/aiEps/fabSection.eps}
    \fi
    \if\whichFig1
        \includegraphics[width=\linewidth]{figures/aiTif/fabSection.tif}
    \fi
    \if\whichFig2
        \includegraphics[width=\linewidth]{figures/aiPdf/fabSection.pdf}
    \fi    	
    \if\whichFig3
    	\includegraphics[width=\linewidth]{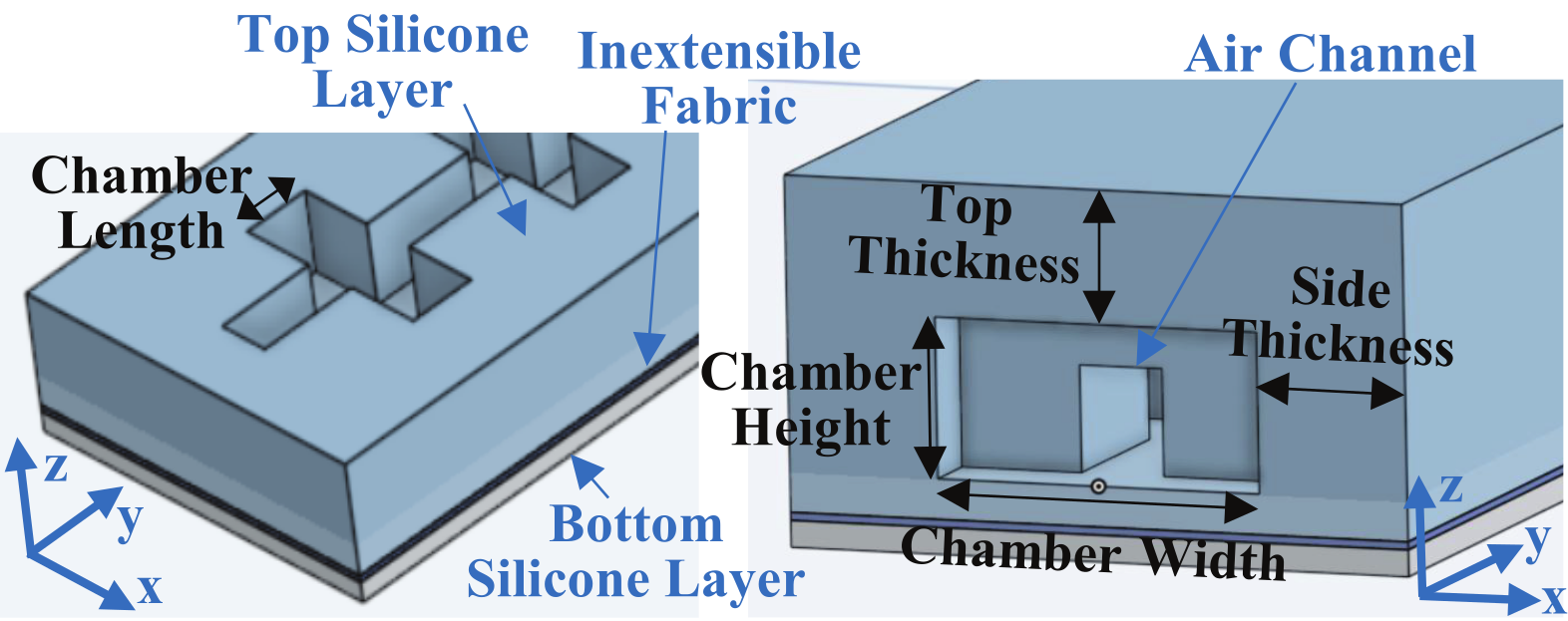}
    \fi    	
    \caption{Section views of the positive CAD model for the standard pneu-net. \emph{Left} X-Y section view. \emph{Right} Z-X section view. Five geometrical parameters are denoted in black, which will be varied.}
    \label{fig.fabSection}
\end{figure}

Fig.~\ref{fig.fabSection} shows section views of the positive CAD model for the standard pneu-net design.
The air chambers are the nodes of the pneumatic network connected by a central air channel.

\begin{figure}[htb]\vspace{-10pt}
    \if\whichFig0
        \includegraphics[width=\linewidth]{figures/aiEps/fabNine3d.eps}
    \fi
    \if\whichFig1
        \includegraphics[width=\linewidth]{figures/aiTif/fabNine3d.tif}
    \fi
    \if\whichFig2
        \includegraphics[width=\linewidth]{figures/aiPdf/fabNine3d.pdf}
    \fi    	
    \if\whichFig3
    	\includegraphics[width=\linewidth]{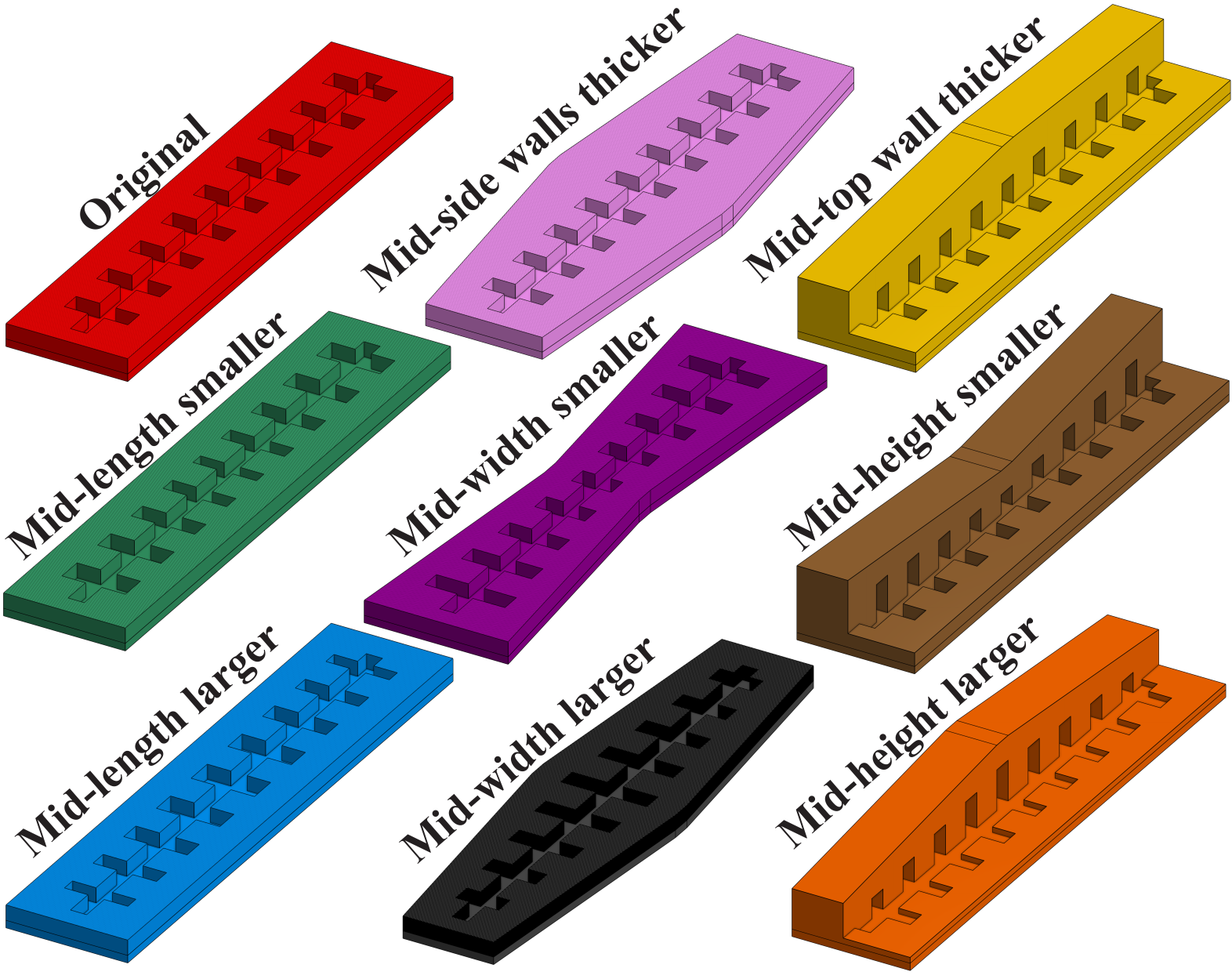}
    \fi    	
    \caption{Nine designs, labeled by how they vary towards the center.}
    \label{fig.fabNine}
    \vspace{-5pt}
\end{figure}

Five parameters that define the main geometry of the pneu-net are denoted in black in Fig.~\ref{fig.fabSection}.
These include air chamber length, air chamber width, air chamber height, silicone side thickness and silicone top thickness.
We vary these parameters along the longitudinal axis of the pneu-net, creating different mechanical codings of behavior.
Fig.~\ref{fig.fabNine} depicts the nine geometrical variations that we created.
The distinct colors will be used as a plot legend in Section~\ref{sec.results}.

For each design, only one of the five parameters is varied.
In the top row, the air chamber volumes remain constant.
The left (red) is the \emph{original} standard design.
The middle (pink) design varies the \emph{side thickness} parameter of the silicone walls, which is increased towards the center and then decreased towards the other end.
The right (yellow) design varies the \emph{top thickness} parameter in the same way.

In the middle row, the air chambers are decreasing in volume towards the center.
The left (green) design allows the \emph{chamber length} to decrease.
The middle (purple) design allows the \emph{chamber width} to decrease.
The right (brown) design allows the \emph{chamber height} to decrease.

In the bottom row, the air chambers are increasing in volume towards the center.
The left (blue) design allows the \emph{chamber length} to increase.
The middle (black) design allows the \emph{chamber width} to increase.
The right (orange) design allows the \emph{chamber height} to increase.

\begin{figure}[htb]
    \if\whichFig0
        \includegraphics[width=\linewidth]{figures/aiEps/fabDimensions.eps}
    \fi
    \if\whichFig1
        \includegraphics[width=\linewidth]{figures/aiTif/fabDimensions.tif}
    \fi
    \if\whichFig2
        \includegraphics[width=\linewidth]{figures/aiPdf/fabDimensions.pdf}
    \fi    	
    \if\whichFig3
    	\includegraphics[width=\linewidth]{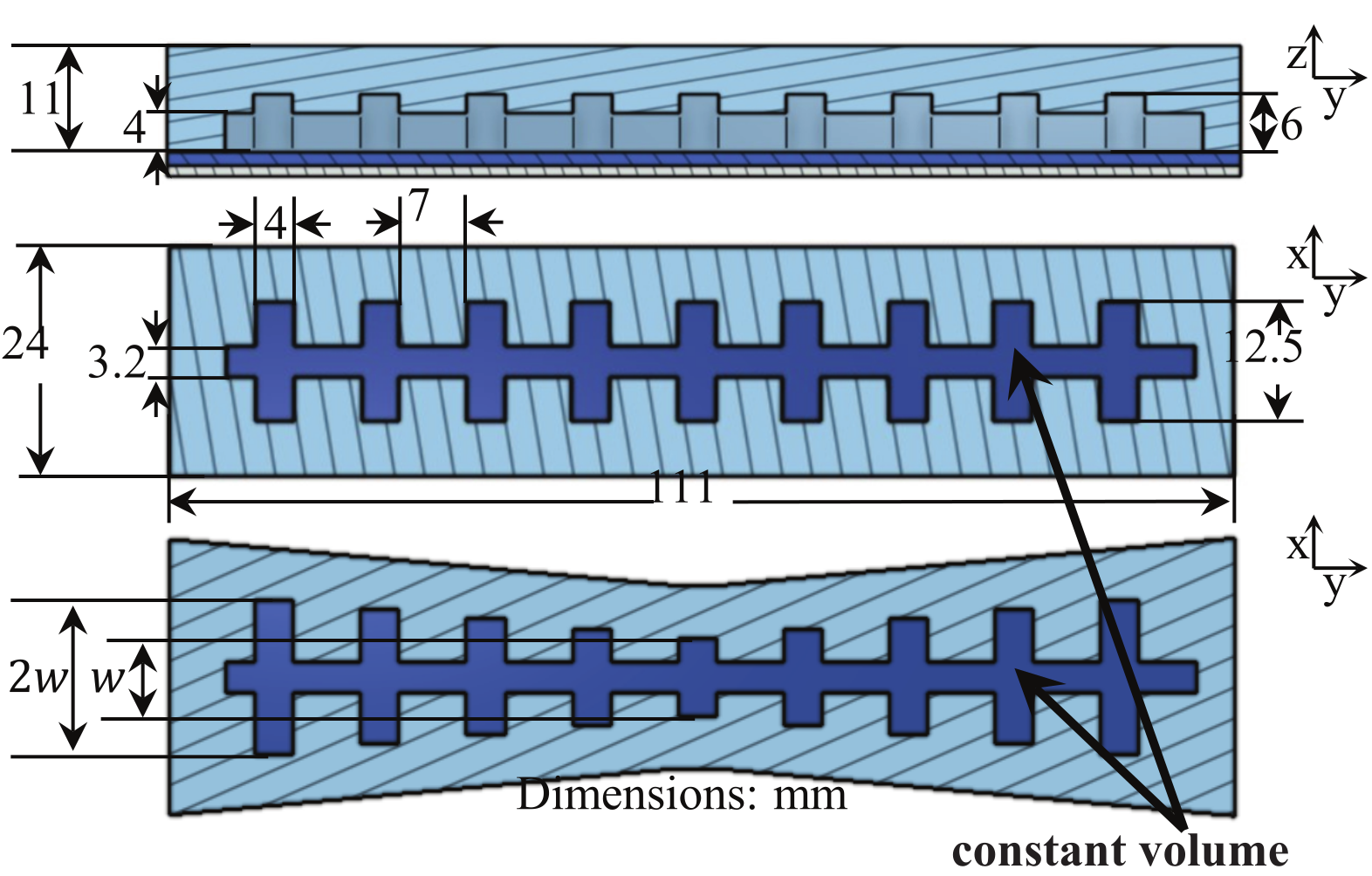}
    \fi    	
    \caption{\emph{(Top and middle)} Side and top section views of \emph{original} design. Dimensions are in mm. \emph{(Bottom)} Top section view of \emph{mid-width smaller} design.}
    \label{fig.fabDimensions}
    \vspace{-10pt}
\end{figure}

In Fig.~\ref{fig.fabDimensions}, the dimensions of the \emph{original} design are shown in the top and middle drawings.
The bottom shows how the \emph{mid-width smaller} design is varied.
The width (x dimension) of each chamber is linearly decreased from the end to the center such that the center width is one-half of the end width.
The width, $w$, is chosen such that the overall internal volume of the actuator is equal to the \emph{original} design.
This process is followed for all designs where air chamber volume varies, such that the internal air volume of the actuator remains constant.
For the designs in the top row of Fig.~\ref{fig.fabNine}, \emph{mid-side walls thicker} and \emph{mid-top wall thicker},
the thickness is varied with a constant slope such that the thickness in the center is 1.5 times the thickness at the ends.

\vspace{-5pt}
\subsection{Fabrication}\label{subsec.fab}  
The pneu-net body is made from Ecoflex 00-50, which has a tensile strength of 315 psi (2.17 N/mm2) and a 980\% elongation at break rating, making it a suitable material for high deformation applications.
The Ecoflex has two parts which are mixed in a 1:1 ratio.
The mixture is then degassed before casting.
The inextensible layer is made from a 100\% cotton plain weave quilting fabric.

\begin{figure}[htb]\vspace{-20pt}
    \subfloat[Pour Mold]{
        \if\whichFig0
            \includegraphics[width=0.49\linewidth]{figures/aiEps/fabPour.eps}
        \fi
        \if\whichFig1
            \includegraphics[width=0.49\linewidth]{figures/aiTif/fabPour.tif}
        \fi
        \if\whichFig2
            \includegraphics[width=0.49\linewidth]{figures/aiPdf/fabPour.pdf}
        \fi    	
	\if\whichFig3
    	    \includegraphics[width=0.49\linewidth]{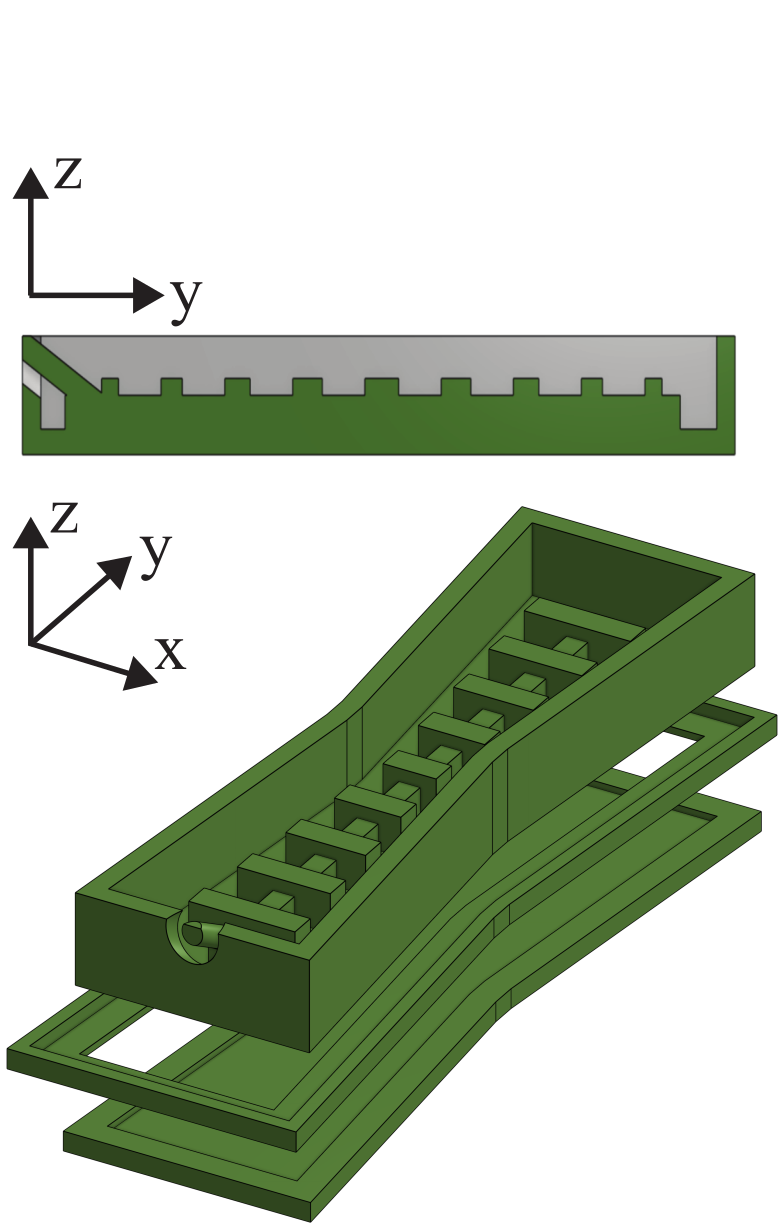}
        \fi    	
        \label{subfig.fabPour}
    }
    \subfloat[Injection Mold]{
        \if\whichFig0
            \includegraphics[width=0.49\linewidth]{figures/aiEps/fabInjection.eps}
        \fi
        \if\whichFig1
            \includegraphics[width=0.49\linewidth]{figures/aiTif/fabInjection.tif}
        \fi
        \if\whichFig2
            \includegraphics[width=0.49\linewidth]{figures/aiPdf/fabInjection.pdf}
        \fi    	
        \if\whichFig3
    	    \includegraphics[width=0.49\linewidth]{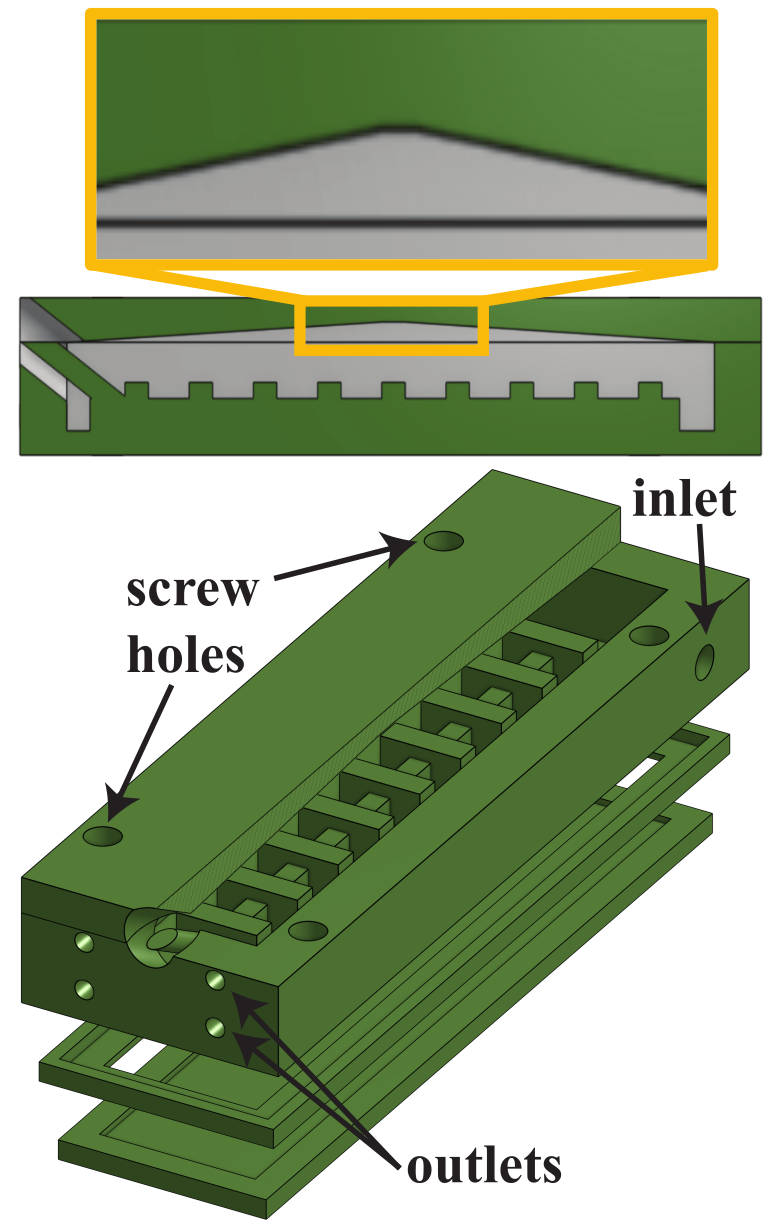}
        \fi    	
        \label{subfig.fabInjection}
    }
    \caption{\protect\subref{subfig.fabPour} Pour mold for the \emph{mid-width smaller} design. From bottom to top: base mold, base frame, top mold, side view of top mold. \protect\subref{subfig.fabInjection} Injection Mold for the \emph{mid-top wall thicker} design. Closeup at the top shows the height-varying nature making injection molding necessary.}
    \label{fig.fabMolds}
\end{figure}

Fig.~\ref{fig.fabMolds} shows CAD models for the molds which are 3D-printed with an Ultimaker 3.
Fig.~\ref{subfig.fabPour} shows the pour mold for the \emph{mid-width smaller} design.
From the bottom to the top, there is the base mold, the frame for the base mold, and the top mold.
The very top shows a side view of the top mold.
Silicone is poured into the bath for the top mold and the shallow bath for the base mold, in the orientation shown, and the silicone is left to settle with gravity.
The limiting layer fabric is placed onto the base mold.
The frame is flipped upside down from the orientation shown here and snapped onto the base and fabric.

Fig.~\ref{subfig.fabInjection} shows the injection mold for the \emph{mid-top wall thicker} design.
Injections molds are needed for the three height-varying designs in the right-most column of Fig.~\ref{fig.fabNine}.
For the non-height-varying designs, pour molds are sufficient since gravity will even out the height along the Z direction.
The side view at the top of Fig.~\ref{subfig.fabInjection} shows a closeup (not to scale) of the height difference, which makes the injection mold necessary.
Note that the base is still a pour mold while the top is an injection mold.
The injection mold is enclosed with screws and placed vertically such that the injection inlet is towards the bottom and the air outlets are at the top.

The rectangular base and frame shown in Fig.~\ref{subfig.fabInjection} are used for the left-most and right-most columns of Fig.~\ref{fig.fabNine}.
For the width-varying designs in the middle column of Fig.~\ref{fig.fabNine}, the base and frame are customized to the shape of the design, as shown in Fig.~\ref{subfig.fabPour}.

\begin{figure}[htb]\vspace{-10pt}
    \if\whichFig0
        \includegraphics[width=\linewidth]{figures/aiEps/fabFlow.eps}
    \fi
    \if\whichFig1
        \includegraphics[width=\linewidth]{figures/aiTif/fabFlow.tif}
    \fi
    \if\whichFig2
        \includegraphics[width=\linewidth]{figures/aiPdf/fabFlow.pdf}
    \fi    	
    \if\whichFig3
    	\includegraphics[width=\linewidth]{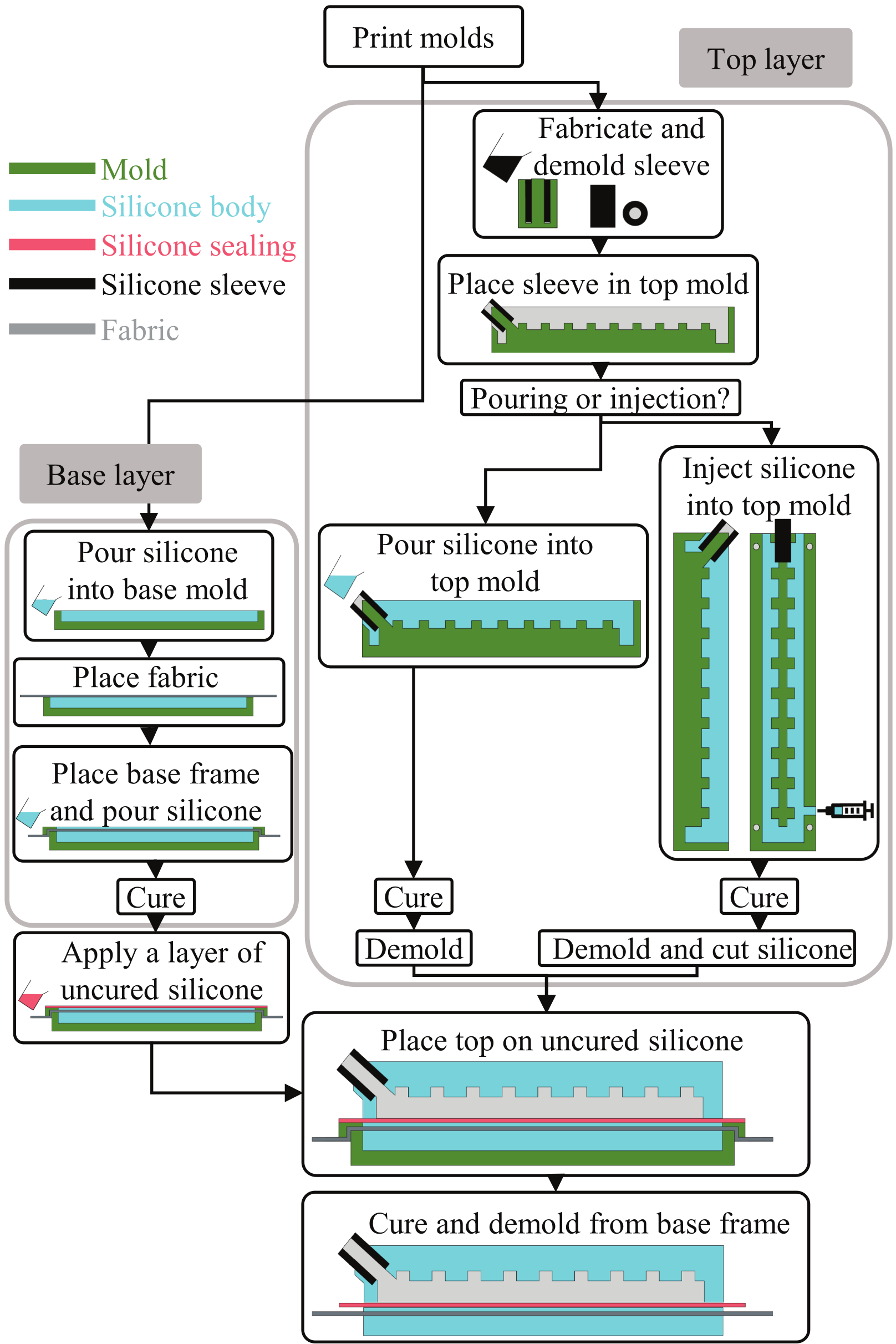}
    \fi    	
    \vspace{-20pt}
    \caption{Flow chart of the fabrication process for both pour and injection molding. Each material is illustrated with a specific color, as denoted by the legend (top left).}
    \label{fig.fabFlow}
\end{figure}

Fig.~\ref{fig.fabFlow} is a flowchart of the fabrication process.
On the left, the base is fabricated.
Silicone is mixed, degassed, and poured into the base mold.
The fabric is placed on top of the poured mixture.
The base frame is snapped on, and another thin layer of silicone is poured.
The base is then left to cure.

On the right of Fig.~\ref{fig.fabFlow}, the top is fabricated.
First, a sleeve is fabricated from Dragon Skin 10 Slow, a stiffer silicone, mixed and degassed in the same manner.
The sleeve is the adapter for the pneumatic control inlet.
The stiffer silicone prevents unwanted expansion in this tube.
For the pour mold, the Ecoflex 00-50 is mixed, degassed, and poured.
For the injection mold, the Ecoflex 00-50 is mixed, degassed, and poured into a syringe.
The mold is placed vertically, and the syringe is injected into the inlet at the bottom.

After the top and base are cured, the top is demolded, and the base is left in its mold and frame.
For the injection mold, the extra silicone from the inlets and outlets are trimmed.
Ecoflex 00-50 is mixed for the sealing layer, and a pigment is added corresponding with the colors shown in Fig.~\ref{fig.fabNine}.
Pigmentation is an optional step, which aids in manually identifying the actuators after fabrication.
The mixture is degassed and poured as a thin film onto the base.
The top is placed on the uncured layer within the perimeter of the base frame.
The base frame serves many purposes:
1) It provides a shallow bath for the thin layer of silicone that is first poured on top of the fabric to embed it completely.
2) It provides a visual guide for placing the top on the base.
3) It controls the exact surface area of contact between the limiting layer and the main body, vital for repeatable behavior when actuated.
After the sealing layer cures, the actuator is demolded from the base mold and frame.


\vspace{-10pt}
\subsection{Finite Element Modeling}\label{subsec.femGeneral}  
FEMs of each of the nine designs are simulated with Abaqus/Standard (Dassault Syst\`{e}mes).
Positive CAD models made from Onshape are imported into the FEM.
The inextensible fabric layer is modeled as isotropic and linearly elastic, with Young's Modulus = 6.5 GPa and Poisson's ratio = 0.2~\cite{Polygerinos2013-IROS}.
The adapter for the air inlet made from stiffer silicone is not modeled.

The Ecoflex 00-50 is modeled as an isotropic hyperelastic solid.
A third-order Yeoh model is fit from stress-strain material testing.
The most common models for stress-strain fits of incompressible materials are Mooney-Rivlin, Neo-Hookean, Yeoh, and Ogden.
Yeoh and Ogden are recommended for strains~$>$~400\%~\cite{marckmann2006comparison},~\cite{bhashyam2002ansys}; therefore, these models are best for silicone rubbers with no reinforcement~\cite{Xavier2021-AdvIntellSyst}, such as the Ecoflex 00-50 that we use.
Ideally, stress-strain data is generated along multiple axes of tension.
If only uniaxial testing is performed, Yeoh will perform better than Ogden~\cite{Xavier2021-AdvIntellSyst},~\cite{shahzad2015mechanical}.
Models for various silicones are reviewed in~\cite{Xavier2021-AdvIntellSyst}.
Out of the models listed for Ecoflex 00-50, all were generated from uniaxial tests; hence Yeoh is more appropriate than Ogden.
Of the Yeoh models, ~\cite{Kulkarni2015-Masters} is the only model fit on data past 400\% strain; indeed, it fits up to 800\% strain.
When fully pressurized, our actuators exhibit strain above 500\%.
(The strain value is obtained by examining the Max. Principal Strain invariants for each element, and then taking the maximum across all elements. Note that it is necessary to request the nominal (engineering) strain field rather than the default logarithmic (true) strain field, and that for multiaxial strain, the conversion between the two is nontrivial~\cite{abqStrain-manual}.)
Models can lead to error if used outside of the deformation range on which they were fit~\cite{marckmann2006comparison}; therefore, it is crucial that the model we use~\cite{Kulkarni2015-Masters} was fit above 500\%.
The coefficients for this model are: $C_{10} = 1.9 \times 10^2$, $C_{20} = 9 \times 10^{-4}$, and $C_{30} = -4.75 \times 10^{-6}$.

After material modeling, the simulation is set up in a standard fashion.
All components are merged.
The merged part is meshed with a global seed size of 2.5mm and no local seeding.
Tetrahedral elements are chosen for greater accuracy.
Free meshing is used for resolution in boundary regions.
The geometric order of node placement is quadratic, and the mesh formulation is hybrid.
An encastre boundary condition is placed on the actuator's fixed end-face. 
Uniform pressure is applied to all internal surfaces as a ramp load from 0 to 50 kPa.

\vspace{-5pt}
\subsection{Data Collection and Control}\label{subsec.data}  
The pneu-nets are actuated with a custom-built pneumatic pump station.
Three positive pressure pumps (Parker BTC-IIS 01620-10) are connected in series to increase maximum pressure.
The pressure source is routed to four air outlets, each connected to a pressure sensor (Honeywell ASDXAVX100PGAA5) and then a solenoid valve.
Pressure control is implemented in MATLAB Simulink, interfacing with the pump station through an Arduino mega 2560.
The pressure readings provide real-time feedback, sampled at 400 Hz.
Since each pressure sensor is located at the input to the actuator instead of the output, its readings are not dampened by the system, so a 5 Hz low pass filter is incorporated as a transfer function into the feedback loop before the pressure is read.
A PWM controller is implemented.
The valves are used in pairs, one for pressure input and the other for an exhaust.
In the \emph{on} state, the pressure input valve is opened, and the exhaust valve is closed.
In the \emph{off} state, the input valve is closed, and the exhaust valve is opened.
The pumps are always run at full capacity, and the exhaust valve ensures the pumps are not damaged.

\begin{figure}[htb]\vspace{-20pt}
    \subfloat[Test Rig]{
        \if\whichFig0
            \includegraphics[width=0.47\linewidth]{figures/aiEps/dataRig.eps}
        \fi
        \if\whichFig1
            \includegraphics[width=0.47\linewidth]{figures/aiTif/dataRig.tif}
        \fi
        \if\whichFig2
            \includegraphics[width=0.47\linewidth]{figures/aiPdf/dataRig.pdf}
        \fi    	
        \if\whichFig3
    	    \includegraphics[width=0.47\linewidth]{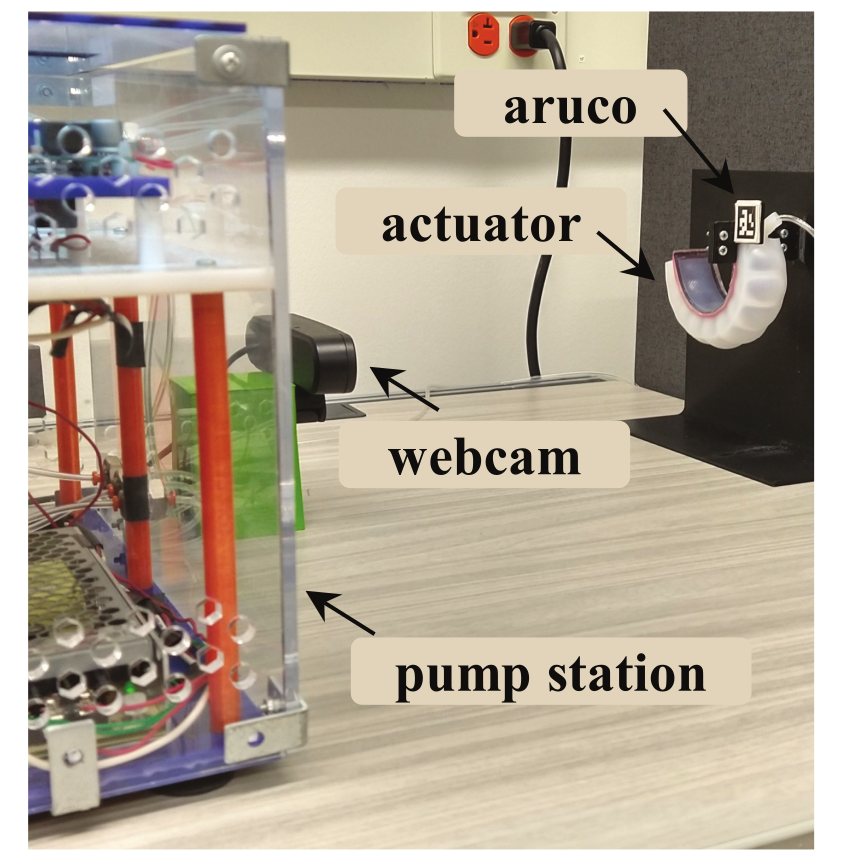}
        \fi    	
	\label{subfig.dataRig}
    }
    \subfloat[DeepLabCut]{
        \if\whichFig0
            \includegraphics[width=0.47\linewidth]{figures/aiEps/dataDlc.eps}
        \fi
        \if\whichFig1
            \includegraphics[width=0.47\linewidth]{figures/aiTif/dataDlc.tif}
        \fi
        \if\whichFig2
            \includegraphics[width=0.47\linewidth]{figures/aiPdf/dataDlc.pdf}
        \fi    	
        \if\whichFig3
    	    \includegraphics[width=0.47\linewidth]{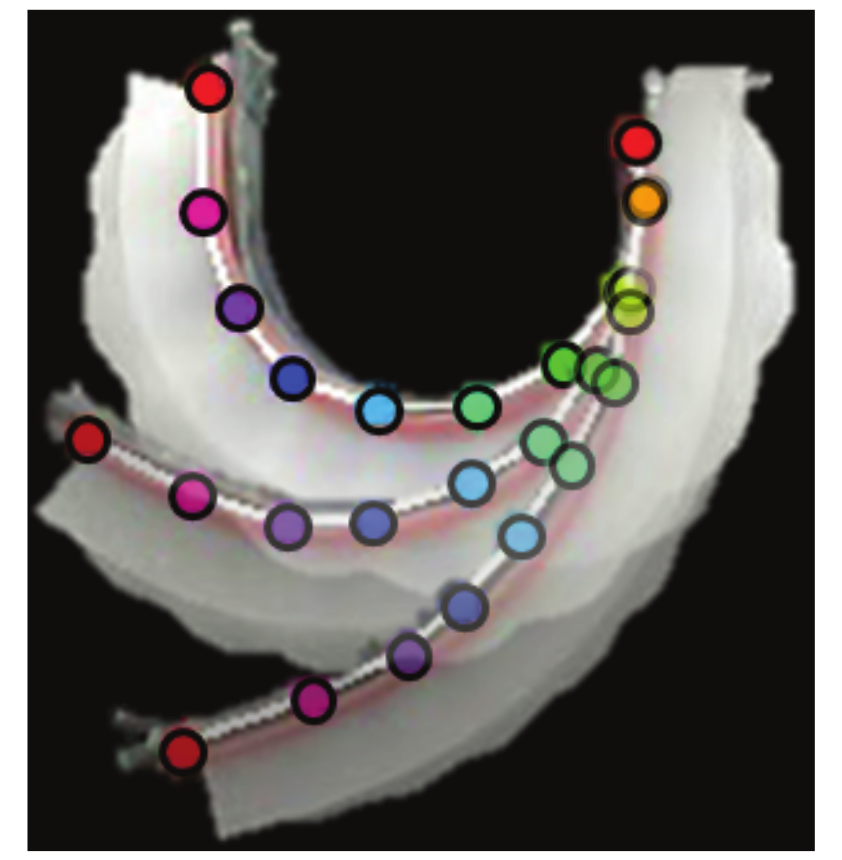}
        \fi    	
	\label{subfig.dataDlc}
    }
    \caption{\protect\subref{subfig.dataRig} Experimental setup showing the actuator being controlled by the pump station while the webcam collects an image stream. \protect\subref{subfig.dataDlc} Three overlaid images of the actuator at different pressures, showing the DLC-tracked points.}
    \label{fig.data}
\end{figure}

Fig.~\ref{subfig.dataRig} shows the experimental setup.
The actuator is being pneumatically controlled.
The webcam is collecting a stream of images at 30 fps, time-synchronized with the pressure readings.

A markerless motion tracking model is trained using the DLC library~\cite{Mathis2018-NatNeurosci} to track the behavior of the actuator.
Fig.~\ref{subfig.dataDlc} shows three image frames of the actuator at different pressure values overlaid with some transparency.
The colored points show the DLC predictions.
Although DLC is typically used for tracking animal body parts, its transfer learning framework allows us to incorporate labeled training data from our soft actuators.
Our training data consists of 30 frames from each actuator with image frames over the full range of input pressures.
The bending angle is calculated between the vector connecting the first two points (the fixed end) and the vector connecting the last two points (the moving end).
This method provides angle measurements with high temporal resolution.

\section{Experiments and Results}\label{sec.results}

\subsection{FEM Results}\label{subsec.femResults}  

\begin{figure}[htb]\vspace{-10pt}
    \if\whichFig0
        \includegraphics[width=\linewidth]{figures/aiEps/staticFemAngleVsPressure.eps}
    \fi
    \if\whichFig1
        \includegraphics[width=\linewidth]{figures/aiTif/staticFemAngleVsPressure.tif}
    \fi
    \if\whichFig2
        \includegraphics[width=\linewidth]{figures/aiPdf/staticFemAngleVsPressure.pdf}
    \fi    	
    \if\whichFig3
    	\includegraphics[width=\linewidth]{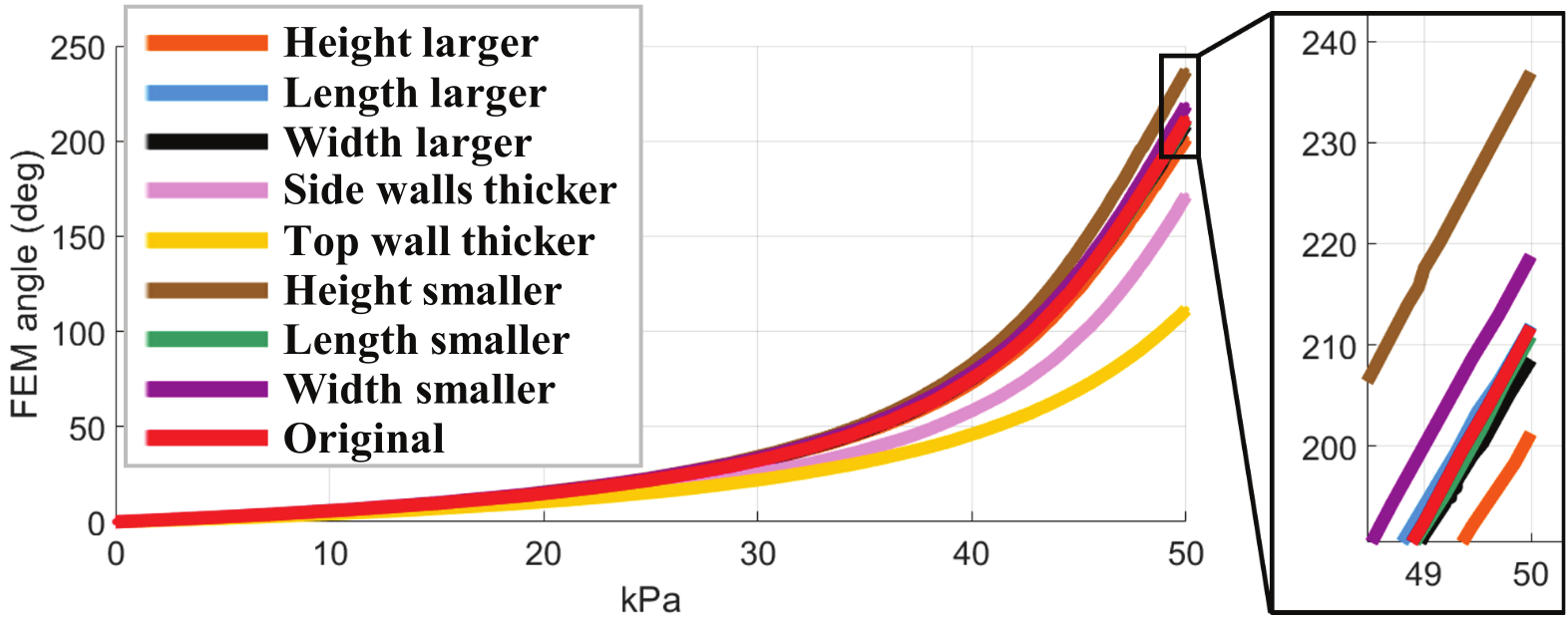}
    \fi    	
    \caption{Angle vs. pressure curves from the FEM models for all nine designs.}
    \label{fig.staticFemAngleVsPressure}
\end{figure}

For each increment of the internal pressure ramp load applied to the actuator, the bending angle is calculated between a vector normal to the fixed end-face of the actuator and a vector normal to the moving end-face.
Fig.~\ref{fig.staticFemAngleVsPressure} shows the angle vs. pressure curves for each of the nine designs.
Note that the colors in the legend correspond to the colors in Fig.~\ref{fig.fabNine}, which will remain consistent throughout all figures.
The closeup at the right end shows the bending angles reached at 50 kPa.

Certain trends can be observed between \emph{sub-families} of designs.
It can be observed that for the designs whose wall thickness increases in the center (pink and yellow), the bending angle is significantly lower.
For the other six designs, where wall thickness stays constant and air chamber volume varies, the bending angles reached are higher.
Out of these six, it can be observed that for the width- and height-varying designs, a smaller volume in the center produces a higher bending angle, while a larger volume in the center produces a lower bending angle.
This trend is not observed for the length-varying designs.

We subsequently compare constant-angle FEM trials for the six designs with the highest bending angle at 50 kPa: \emph{mid-height smaller} (brown), \emph{mid-width smaller} (purple), \emph{mid-length smaller} (green), \emph{original} (red), \emph{mid-length larger} (blue), and \emph{mid-width larger} (black).
Out of these six, the design with the smallest bending angle is \emph{mid-width larger} (black), at 208$^{\circ}$.
For each design simulation, we query the increment in the 0 to 50 kPa ramp load that corresponds to 208$^{\circ}$.
In this way, we can compare the stress distributions of these six designs at the same angle.
We ignore the three designs with the lowest bending angle so that we are not limited to too low of a constant angle for the comparison.

\begin{figure}[htb]\vspace{-5pt}
    \if\whichFig0
        \includegraphics[width=\linewidth]{figures/aiEps/femBasic.eps}
    \fi
    \if\whichFig1
        \includegraphics[width=\linewidth]{figures/aiTif/femBasic.tif}
    \fi
    \if\whichFig2
        \includegraphics[width=\linewidth]{figures/aiPdf/femBasic.pdf}
    \fi    	
    \if\whichFig3
    	\includegraphics[width=\linewidth]{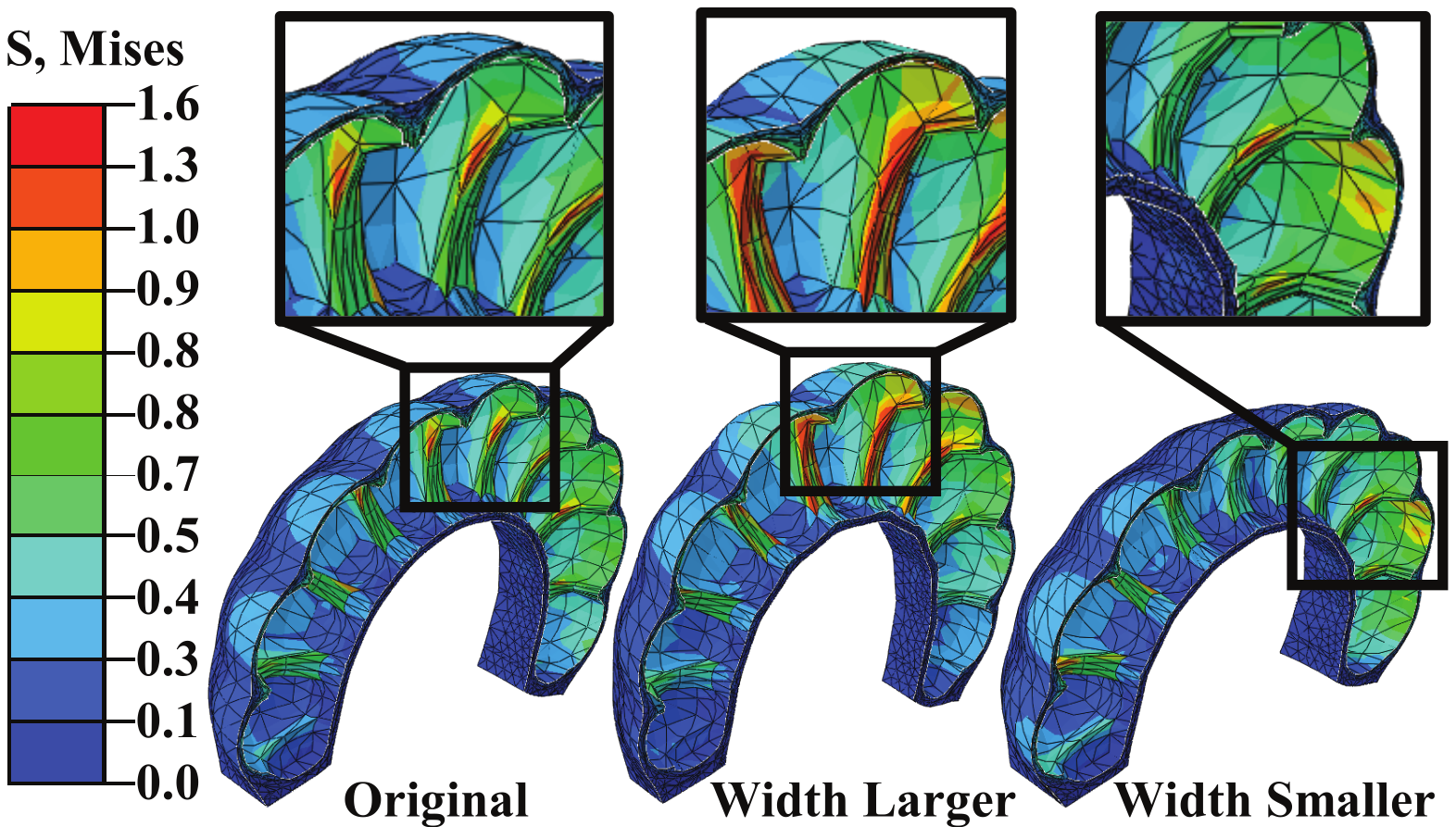}
    \fi    	
    \vspace{-20pt}
    \caption{Von Mises stress distributions of the FEM models for three designs: Original, Width Larger, and Width Smaller. Units are in MPa. Y-Z sectional views are shown, with closeups above highlighting the maximum stress within the interior walls and how the design choice affects the stress level.}
    \label{fig.femBasic}\vspace{-5pt}
\end{figure}

Fig.~\ref{fig.femBasic} shows y-z section views of the FEM for three of these six designs: \emph{original}, \emph{mid-width larger}, and \emph{mid-width smaller}. 
They are all shown pressurized to an increment corresponding to 208$^{\circ}$.
The color bar shows the equivalent von Mises stress in units of MPa.
Note that the inextensible layers are removed from the diagrams because their stresses are an order of magnitude higher.
We are interested in the color scales of the extensible layer to compare the differences between geometries.
In all three designs, the maximum stress (shown in red) occurs in the interior walls, as expected for the sPN design.
For the \emph{original} design, the stress is highest in the center because the center air chambers expand slightly more than the distal ones.
For \emph{mid-width larger}, the stress on the middle air chambers becomes more prominent and the maximum stress increases.
For \emph{mid-width smaller}, the stress on the middle air chambers becomes less prominent, and is redistributed out to the distal chambers.

\begin{figure}[htb]
    \if\whichFig0
        \includegraphics[width=\linewidth]{figures/aiEps/staticStressDistributionsAll.eps}
    \fi
    \if\whichFig1
        \includegraphics[width=\linewidth]{figures/aiTif/staticStressDistributionsAll.tif}
    \fi
    \if\whichFig2
        \includegraphics[width=\linewidth]{figures/aiPdf/staticStressDistributionsAll.pdf}
    \fi    	
    \if\whichFig3
    	\includegraphics[width=\linewidth]{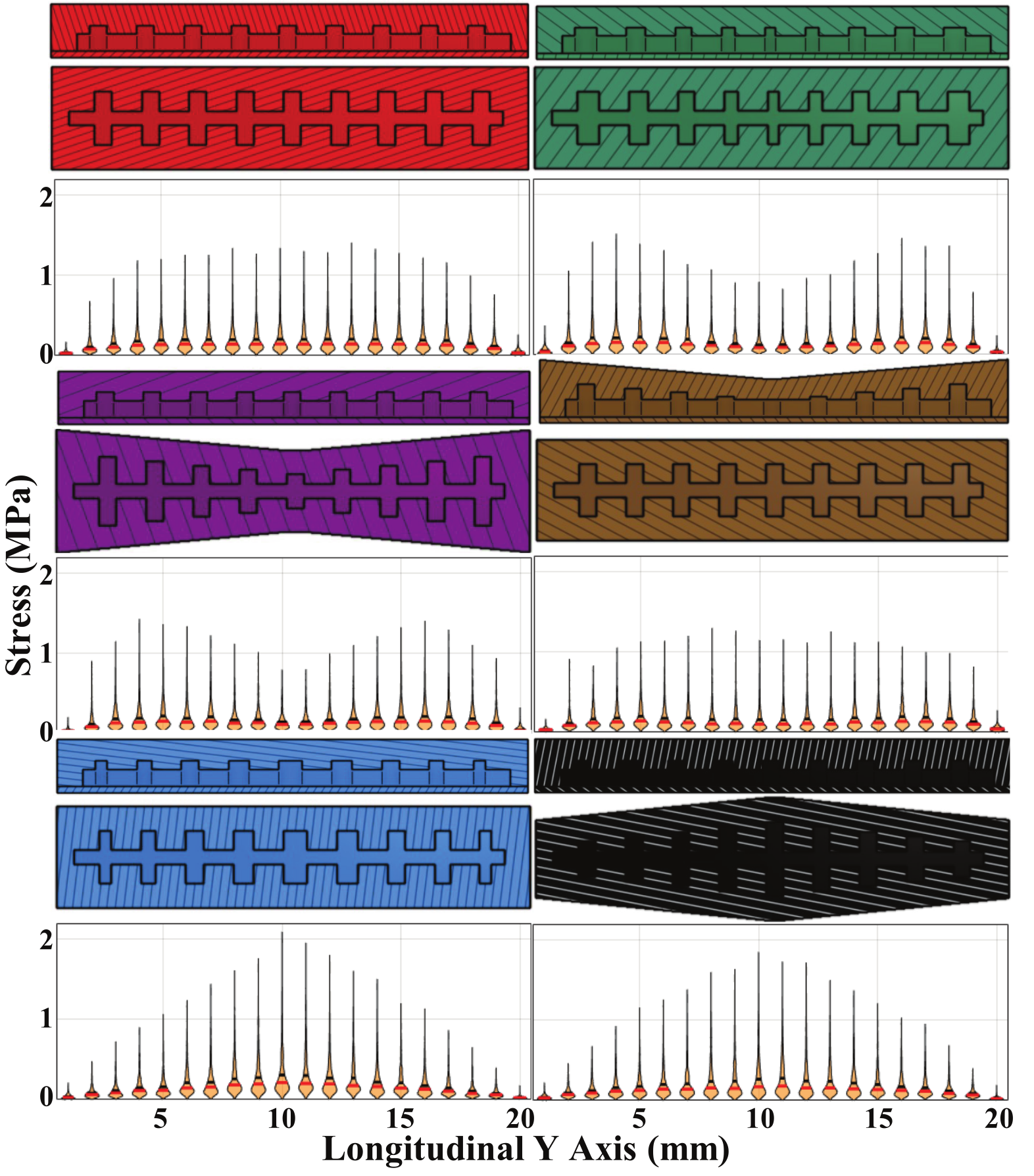}
    \fi    	
    \vspace{-20pt}
    \caption{Stress distributions for the six designs that reach the highest bending angle, reported at a bending angle of 208\%. Each graph shows violin distributions for 20 cross-sectional slices along the longitudinal axis of each actuator. The Y-Z and Y-X section views of each design are shown above the violins such that the longitudinal (y) axes of the section views correspond with the horizontal axis of the violin plot.}
    \label{fig.staticStressDistributionsAll}
    \vspace{-10pt}
\end{figure}

Fig.~\ref{fig.staticStressDistributionsAll} is a more quantitative analysis in post-processing of the stress patterns depicted in Fig.~\ref{fig.femBasic}.
Here, we examine all six designs with the highest bending angle, compared at 208$^{\circ}$.
For each design, we evenly divide the geometry into 20 cross-sectional slices normal to the longitudinal (Y) axis.
For the graph of each design, we portray 20 violin plots, one for each of the cross sections.
Note that for each design, we show the y-z and y-x section views above the violins such that the longitudinal (Y) axes of the section views correspond with the horizontal axis of the violin plot.
Each violin plot shows the distribution of von Mises stress values for the finite elements within that cross section.
Again, the inextensible layers are removed from the scales since they are an order of magnitude higher, and we are interested in comparing the differences in geometries of the extensible parts.
All six graphs are shown on the same scale of 0 to 2 MPa for comparison.

For the original design, the stress stays relatively constant throughout the middle slices and decreases at the boundaries.
All of the designs similarly show this pattern of stress decrease at the boundaries.
Factoring out this boundary effect, for the width- and length-varying design sub-families (purple, black, green, and blue), the volume of the air chambers is positively correlated with the amount of stress in its surrounding silicone walls.
For the height-varying sub-family (brown), this correlation is not observed.

\vspace{-10pt}
\subsection{Static Analysis}\label{subsec.static}  

\begin{figure}[htb]\vspace{-10pt}
    \if\whichFig0
        \includegraphics[width=\linewidth]{figures/aiEps/staticAngleVsPressureMult.eps}
    \fi
    \if\whichFig1
        \includegraphics[width=\linewidth]{figures/aiTif/staticAngleVsPressureMult.tif}
    \fi
    \if\whichFig2
        \includegraphics[width=\linewidth]{figures/aiPdf/staticAngleVsPressureMult.pdf}
    \fi    	
    \if\whichFig3
    	\includegraphics[width=\linewidth]{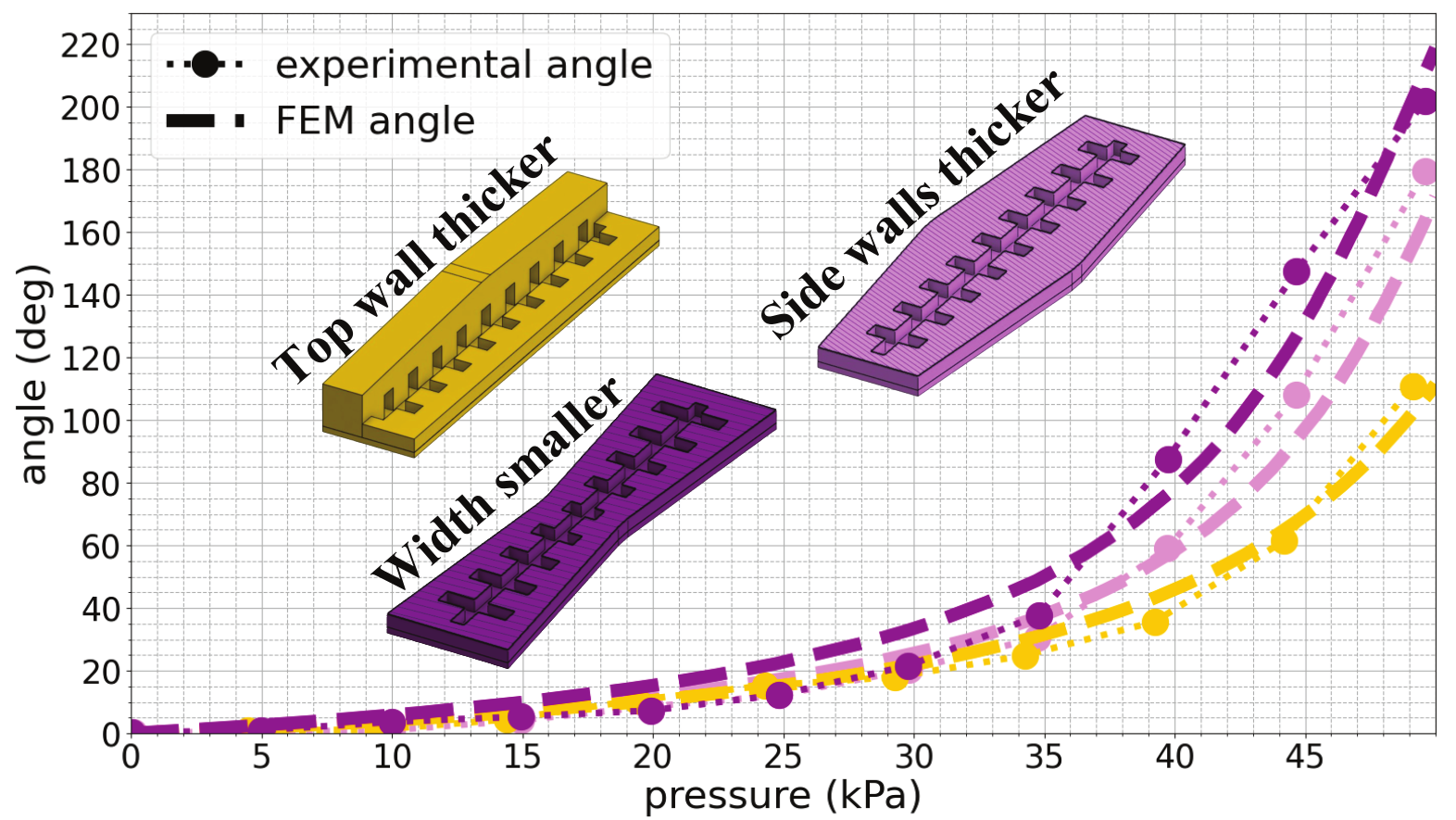}
    \fi    	
    \caption{Angle vs. pressure curves for three designs, comparing FEM and experimental data.}
    \label{fig.staticAngleVsPressureMult}
    \vspace{-5pt}
\end{figure}

Here we validate our FEM models with experimental data.
Fig.~\ref{fig.staticAngleVsPressureMult} shows bending angle vs. pressure curves for three out of the nine designs: \emph{mid-width smaller} (purple), \emph{mid-side walls thicker} (pink), and \emph{mid-top wall thicker} (yellow).
The thick dashed lines are the FEM curves, as shown in Fig.~\ref{fig.staticFemAngleVsPressure}.

The circular points (connected with thin dashed lines) in Fig.~\ref{fig.staticAngleVsPressureMult} are experimental data.
In order to acquire data for the static response experiments, each actuator was pneumatically controlled with a staircase reference function.
The staircase consisted of step functions increasing in 5 kPa increments, from 0 to 50 kPa.
At each 5 kPa step, the pressure is held constant for some time so that the actuator can reach steady state.
We measure the bending angle using our vision model at each pressure step once the actuator has reached steady state.
Thus, we can measure the actuator's static response to each given pressure, which can then be compared to the static FEM model.

We can qualitatively observe in Fig.~\ref{fig.staticAngleVsPressureMult} that the FEM and experimental data have a strong correlation.
It also is seen that the order of bending angle values is preserved: \emph{mid-width smaller} (purple) is the highest, and \emph{mid-top wall thicker} (yellow) is the lowest.
The accuracies of the FEM models are quantitatively validated in Table~\ref{table.femVsExp}.
The error is calculated between the FEM and experimental data for these three designs.
The normalized RMSE (NRMSE) is $\le$ 5\% for all three designs.
Note that these experiments are conducted on new actuators on their first trial of pressurization after fabrication.
Refer to our work in~\cite{Libby2023-ISMR}, for an analysis on how this error increases with fatigue.

\begin{table}[htb]\vspace{-10pt}
	\renewcommand{\arraystretch}{1.3} 
	\caption[]{Model Validation against Experimental Data}\vspace{-5pt}
	\centering
	\begin{tabular}{ l | l | l}
		\hline
		\textbf{Design} & \textbf{RMSE} & \textbf{NRMSE} \\
		\hline
		\hline
		Side walls thicker & 7.29$^{\circ}$ & 4.0\% \\
		\hline
		Top wall thicker & 4.35$^{\circ}$ & 4.0\% \\
		\hline
		Width smaller & 9.99$^{\circ}$ & 5.0\% \\
		\hline
	\end{tabular}
	\label{table.femVsExp} 
\end{table}

\begin{figure}[htb]\vspace{-10pt}
    \if\whichFig0
        \includegraphics[width=\linewidth]{figures/aiEps/staticAngleSideBySides.eps}
    \fi
    \if\whichFig1
        \includegraphics[width=\linewidth]{figures/aiTif/staticAngleSideBySides.tif}
    \fi
    \if\whichFig2
        \includegraphics[width=\linewidth]{figures/aiPdf/staticAngleSideBySides.pdf}
    \fi    	
    \if\whichFig3
    	\includegraphics[width=\linewidth]{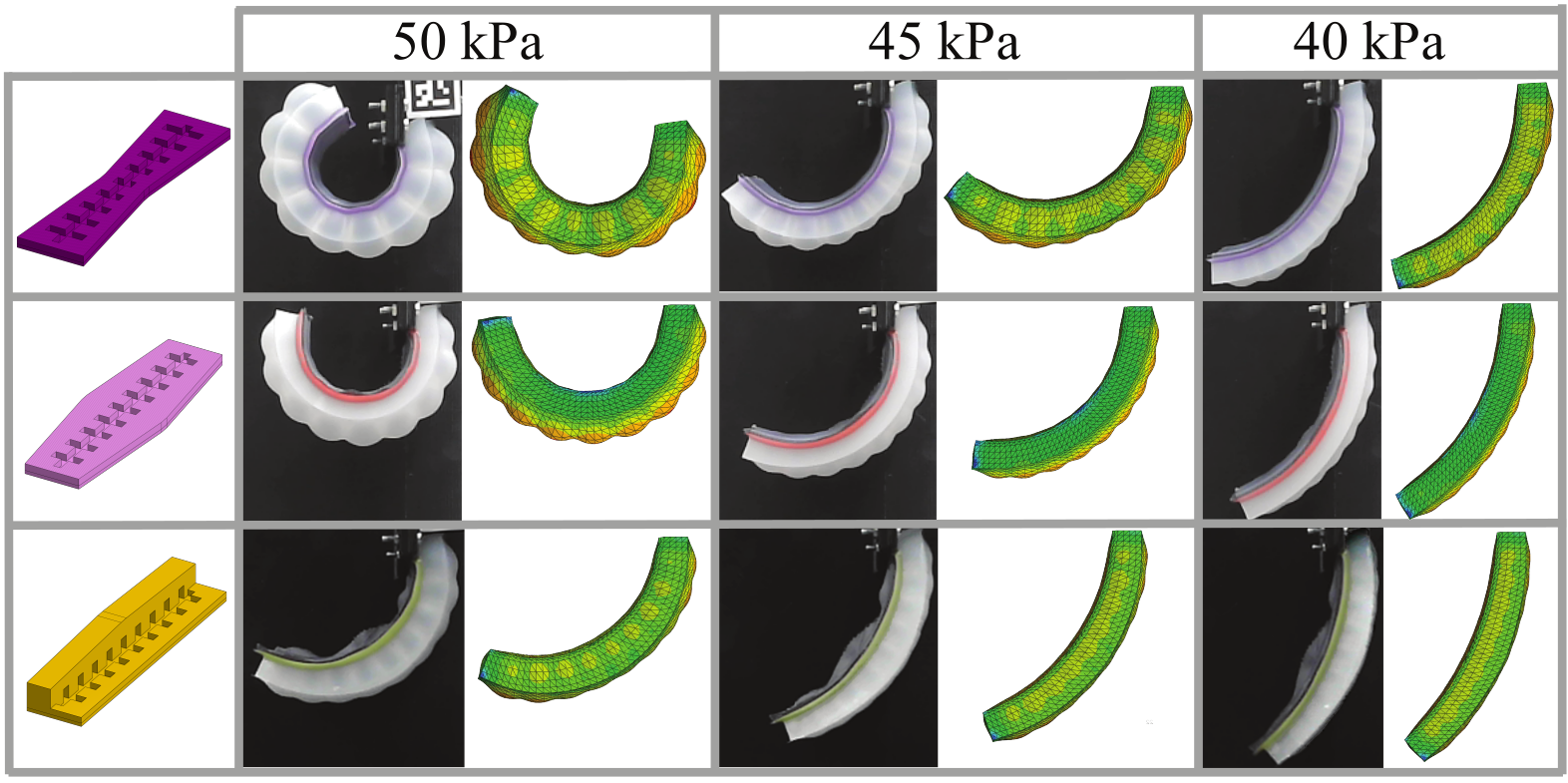}
    \fi    	
    \caption{Side-by-side comparisons of experimental webcam images and simulation images for three designs at three static pressure values.}
    \label{fig.staticAngleSideBySides}
    \vspace{-10pt}
\end{figure}

Fig.~\ref{fig.staticAngleSideBySides} visually compares the FEM vs. experimental data for the same three designs.
For the final three staircase steps of 40, 45, and 50 kPa, webcam images are shown for the experimental data, and simulation images are shown for the FEM data.
This visual comparison allows the reader to qualitatively appreciate the similarity in the bending angle parameter as well as the continuum of passive degrees of freedom that define the overall shape.

\vspace{-10pt}
\subsection{Dynamic Analysis}\label{subsec.dynamic}  

For dynamic trials, each actuator was pneumatically controlled with a triangle wave reference function.
Trials were conducted at high and low frequencies.
The high frequency was 0.25Hz, which was empirically chosen.
At frequencies $>$ 0.25 Hz, the actuators do not have time to fully deflate when pressurized to a bending angle $>$ 180$^{\circ}$.
The low frequency was chosen to be 0.0625 Hz, $1/4$ of the high frequency.

\begin{figure}[htb]
    \subfloat[Time Series]{
	    \begin{minipage}[b]{.49\hsize}\vspace{-20pt}
                \if\whichFig0
                    \includegraphics[width=\linewidth]{figures/aiEps/dynTime.eps}
                \fi
                \if\whichFig1
                    \includegraphics[width=\linewidth]{figures/aiTif/dynTime.tif}
                \fi
                \if\whichFig2
                    \includegraphics[width=\linewidth]{figures/aiPdf/dynTime.pdf}
                \fi    	   
    		\if\whichFig3
    		    \includegraphics[width=\linewidth]{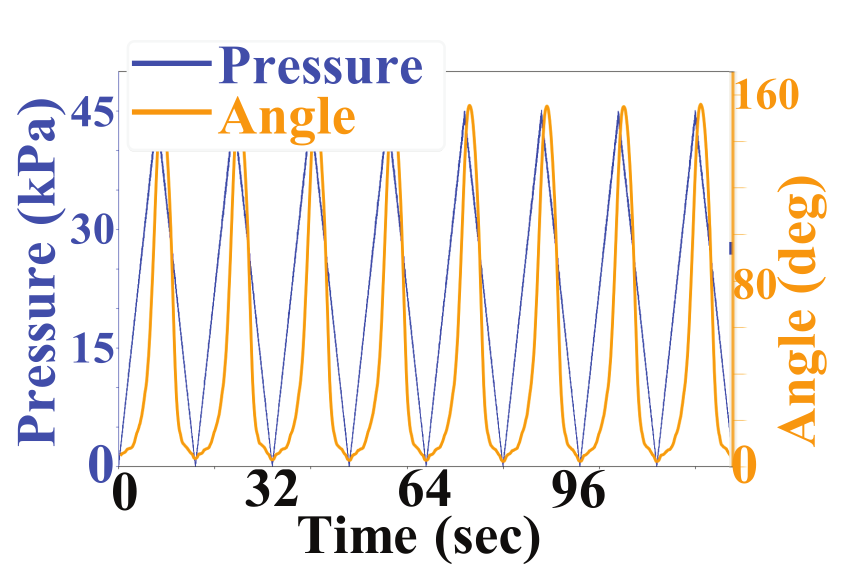}
    		\fi    	         
                \label{subfig.dynTime}
                \vspace{-10pt}
            \end{minipage}%
    }
    \subfloat[Angle vs. Pressure]{
            \begin{minipage}[b]{.49\hsize}\vspace{-20pt}
                \if\whichFig0
                    \includegraphics[width=\linewidth]{figures/aiEps/dynLeafOne.eps}
                \fi
                \if\whichFig1
                    \includegraphics[width=\linewidth]{figures/aiTif/dynLeafOne.tif}
                \fi
                \if\whichFig2
                    \includegraphics[width=\linewidth]{figures/aiPdf/dynLeafOne.pdf}
                \fi    	   
    		\if\whichFig3
    		    \includegraphics[width=\linewidth]{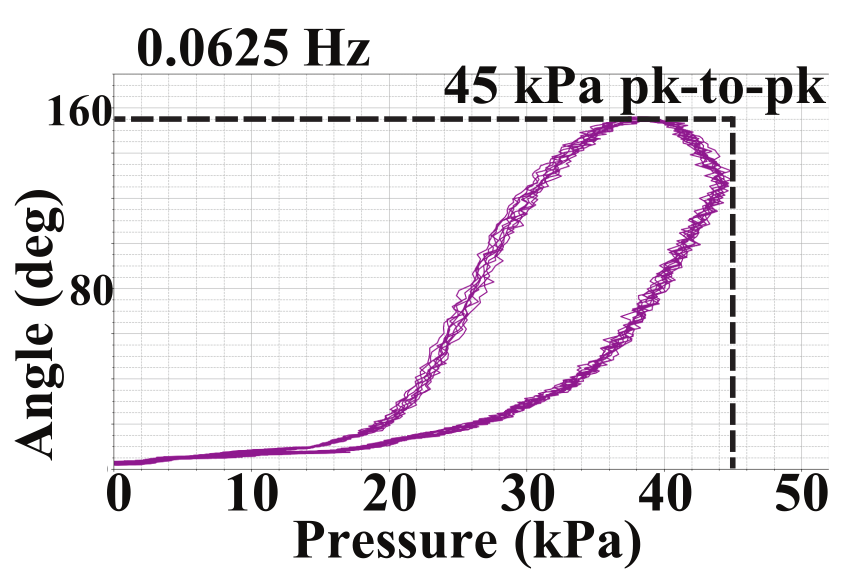}
    		\fi    	                  
                \label{subfig.dynLeafOne}
                \vspace{-10pt}
            \end{minipage}%
    }
    \caption{One trial of a 0.0625Hz triangle wave with a 45 kPa peak-to-peak pressure input. \emph{(Left)} YY plot of the pressure and bending angle readings. Eight of the 12 cycles are shown. The angle readings come from the DLC markerless tracking. \emph{(Right)} The same data showing angle vs. pressure, depicting hysteretic behavior. All 12 triangle cycles are shown here, overlapping with high repeatability.}
    \label{fig.dynOne}
    \vspace{-10pt}
\end{figure}

Fig.~\ref{subfig.dynTime} shows an example trial of a 0.0625 Hz triangle with a 45 kPa peak-to-peak amplitude.
The YY plot shows input pressure vs. time (in blue) and bending angle vs. time (in orange).
The orange bending angle can be observed to closely follow the blue triangle wave.

\fboxrule=0pt 
\fboxsep=0pt 

Note that the measured pressure (in blue) follows the reference triangle wave almost perfectly.
Table~\ref{table.tracking} reports the very low tracking errors of the measured pressure to the reference pressure for all control functions used in experimentation.
For the triangle waves, the errors are reported for 55 kPa peak-to-peak amplitudes, the highest range we use.

\begin{table}[htb]\vspace{-10pt}
	\renewcommand{\arraystretch}{1.3} 
	\caption[]{Tracking Error of Pneumatic Control}\vspace{-5pt}
	\centering
	\begin{tabular}{ l | l | l}
		\hline
		\textbf{Ref. function} & \textbf{RMSE} & \textbf{NRMSE} \\
		\hline
		\hline
		Staircase & 0.4kPa & 0.9\% \\
		\hline
		0.0625 Hz Triangle & 0.4 kPa & 0.7\% \\
		\hline
		0.025 Hz Triangle & 0.5 kPa & 1.0\% \\
		\hline
	\end{tabular}
	\label{table.tracking} 
\end{table}

Fig.~\ref{subfig.dynLeafOne} is a parametric plot of the pressure and angle time series data from Fig.~\ref{subfig.dynTime}.
One loop of the leaf-like perimeter represents one triangle cycle.
Each trial is conducted for 12 cycles, and it can be observed that the 12 leaves overlap with high repeatability, demonstrating the robustness of the various components of our system, including time synchronization, pressure measurements, pneumatic control, and computer vision angle predictions.
Moreover, the overlap demonstrates high repeatability in the dynamic behavior of the actuator.

The vertical dashed line in Fig.~\ref{subfig.dynLeafOne} denotes the maximum pressure, which is 45 kPa (corresponding to the 45 kPa peak-to-peak triangle in Fig.~\ref{subfig.dynTime}).
The horizontal dashed line in Fig.~\ref{subfig.dynLeafOne} denotes the maximum angle, which is just under 160$^{\circ}$ for this trial (corresponding to the peak of the orange curve in Fig.~\ref{subfig.dynTime}).

The cumulative hysteresis is calculated as the area within the leaf.
For many applications, it is desirable to minimize the cumulative hysteresis so that a given input pressure maps to similar bending angles for succeeding inflation and deflation states of the actuator.

The \emph{hysteresis ratio}~\cite{Berman2005-EngStruct,Fedullo1980-AmRevRespirDis,Bachofen1971-JApplPhysiol} is a normalized metric of the cumulative hysteresis, normalized by both the input and the output.
This is the ratio of the area within the leaf to the area of the bounding rectangle.
The bounding rectangle is defined by the axes and the dashed lines in Fig.~\ref{subfig.dynLeafOne}.


In most literature where the hysteresis between different trials is compared, the load is a pressure input or a stress measurement output, and the other variable (input or output) is some measure of strain (elongation or bending).
In cases where the input or the output varies greatly between compared trials, that variable should be normalized before making the hysteresis comparison.
For highly-deformable soft actuators, the bending angle (strain output) can vary greatly, so this normalization is critical.
The hysteresis ratio normalizes both the input and output, so variation in either variable is accounted for.
This normalization technique can be found in many applications ranging from physics~\cite{Chung2013-JPhysD} to civil engineering~\cite{Berman2005-EngStruct} to human physiology~\cite{Fedullo1980-AmRevRespirDis},~\cite{Bachofen1971-JApplPhysiol}.

\begin{figure}[htb]\vspace{-10pt}
    \if\whichFig0
        \includegraphics[width=\linewidth]{figures/aiEps/dynLeavesAll.eps}
    \fi
    \if\whichFig1
        \includegraphics[width=\linewidth]{figures/aiTif/dynLeavesAll.tif}
    \fi
    \if\whichFig2
        \includegraphics[width=\linewidth]{figures/aiPdf/dynLeavesAll.pdf}
    \fi    	
    \if\whichFig3
    	\includegraphics[width=\linewidth]{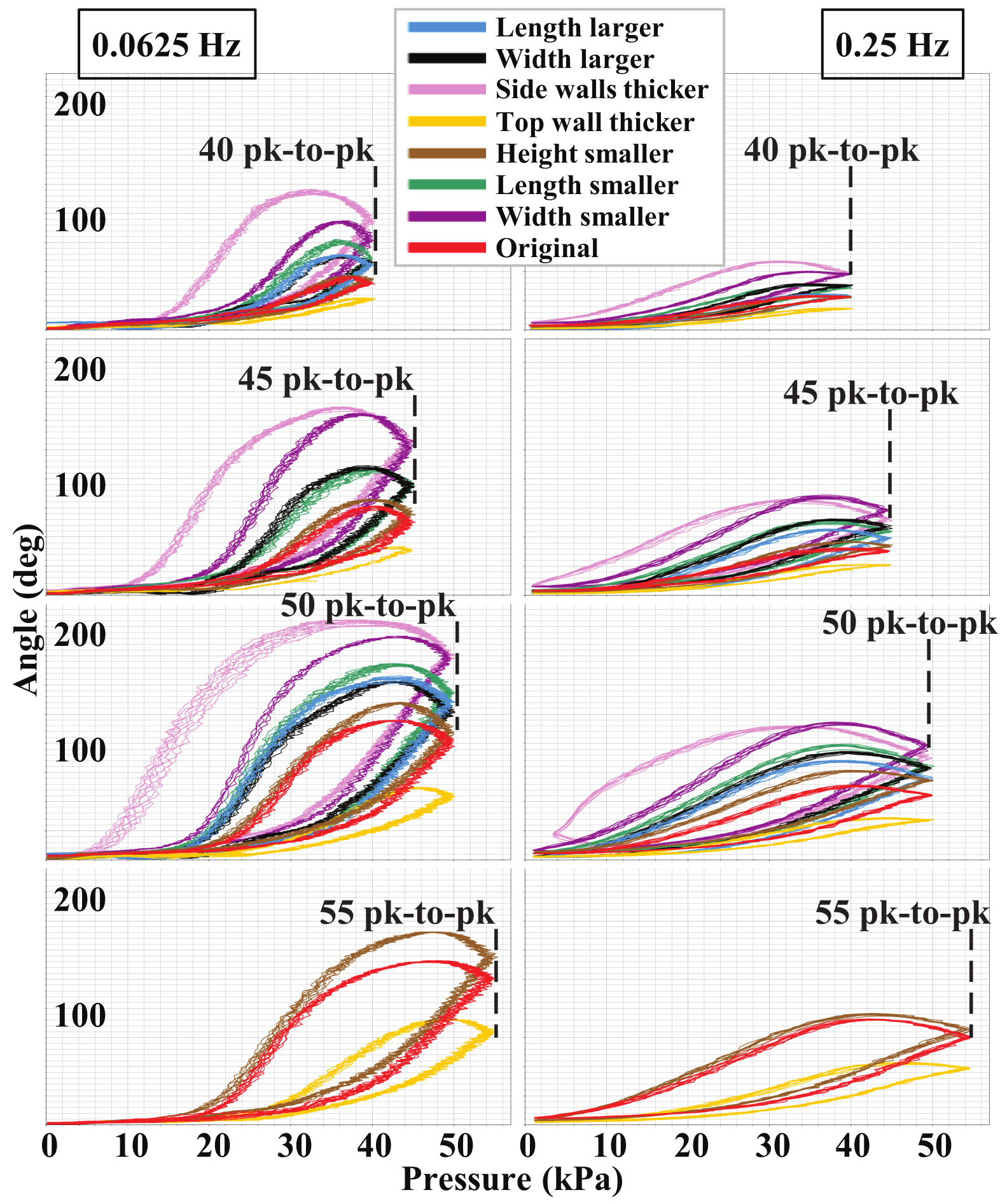}
    \fi    	         
    \vspace{-20pt}
    \caption{Angle vs. pressure for all experimental trials. The left column are trials at 0.0625 Hz. The right column are trials at 0.25 Hz. Each row shows the trials at peak-to-peak pressures of 40, 45, 50, and 55 kPa. The legend shows the colors for each of the eight designs. Each leaf in each plot shows a trial of 12 oscillations for the design of that color.}
    \label{fig.dynLeavesAll}
    \vspace{-10pt}
\end{figure}

Fig.~\ref{fig.dynLeavesAll} shows the hysteresis curves for all dynamic experiments.
Low-frequency trials are in the left column and high-frequency in the right.
The four rows present the 40, 45, 50, and 55 kPa peak-to-peak pressures, respectively.
Each plot is similar to Fig.~\ref{subfig.dynLeafOne}, where the x-axis is pressure, the y-axis is bending angle, and the resulting leaf is created from 12 triangle cycles on that actuator.
Experimental trials are conducted for eight designs, with legend colors consistent with previous figures.
(The ninth \emph{mid-height larger} design was not tested due to experimental limitations.)
So, for example, in the top left plot, all designs are run with a 0.0625 Hz triangle wave and a 40 kPa peak-to-peak pressure.
It can be observed that the design geometry dramatically affects the dynamic response (the maximum angle reached) given the same input pressure.
This trend can be observed in all frequency and pressure combinations.
Note that for the 55 kPa peak-to-peak pressure, we only conduct trials on the subset of designs with the lowest dynamic response so that constant angle comparisons can subsequently be made at a higher angle.
Also note that all trials were conducted on newly fabricated actuators such that the effect of fatigue was minimized.

It can be observed that the design geometry affects the cumulative hysteresis.
For instance, in the top left graph (0.0625 Hz, 45 kPa pk-to-pk), comparing the \emph{mid-width smaller} (purple) and \emph{mid-side walls thicker} (pink) designs, the dynamic response is similar, but the cumulative hysteresis for the pink design is larger.
In this case, there is a clear winner because both actuators are being compared at the same peak-to-peak pressure and the same peak-to-peak bending angle.

\begin{figure}[htb]
    \subfloat[Bar Charts of Dynamic Response and Hysteresis Ratio]{
        \if\whichFig0
            \includegraphics[width=1\linewidth]{figures/aiEps/dynHystBarAll.eps}
        \fi
        \if\whichFig1
            \includegraphics[width=1\linewidth]{figures/aiTif/dynHystBarAll.tif}
        \fi
        \if\whichFig2
            \includegraphics[width=1\linewidth]{figures/aiPdf/dynHystBarAll.pdf}
        \fi    	
    	\if\whichFig3
    	    \includegraphics[width=1\linewidth]{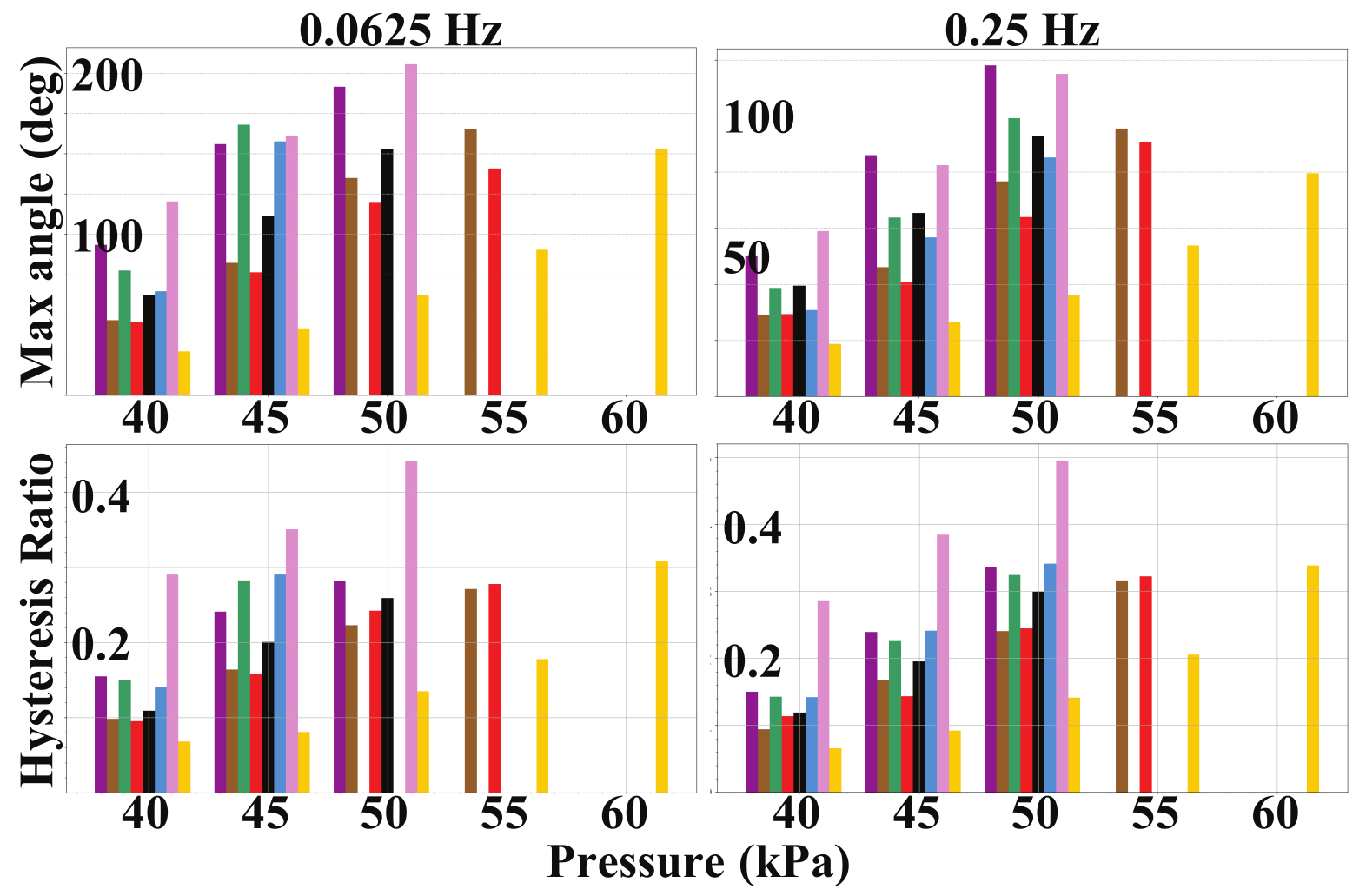}
        \fi    	         
        \label{subfig.dynHystBar}
    }
    \vfill
    \subfloat[Hysteresis Ratio Comparisons at Constant Angle]{
        \if\whichFig0
            \includegraphics[width=1\linewidth]{figures/aiEps/dynHystVsAngleVertBoth.eps}
        \fi
        \if\whichFig1
            \includegraphics[width=1\linewidth]{figures/aiTif/dynHystVsAngleVertBoth.tif}
        \fi
        \if\whichFig2
            \includegraphics[width=1\linewidth]{figures/aiPdf/dynHystVsAngleVertBoth.pdf}
        \fi    	
    	\if\whichFig3
    	    \includegraphics[width=1\linewidth]{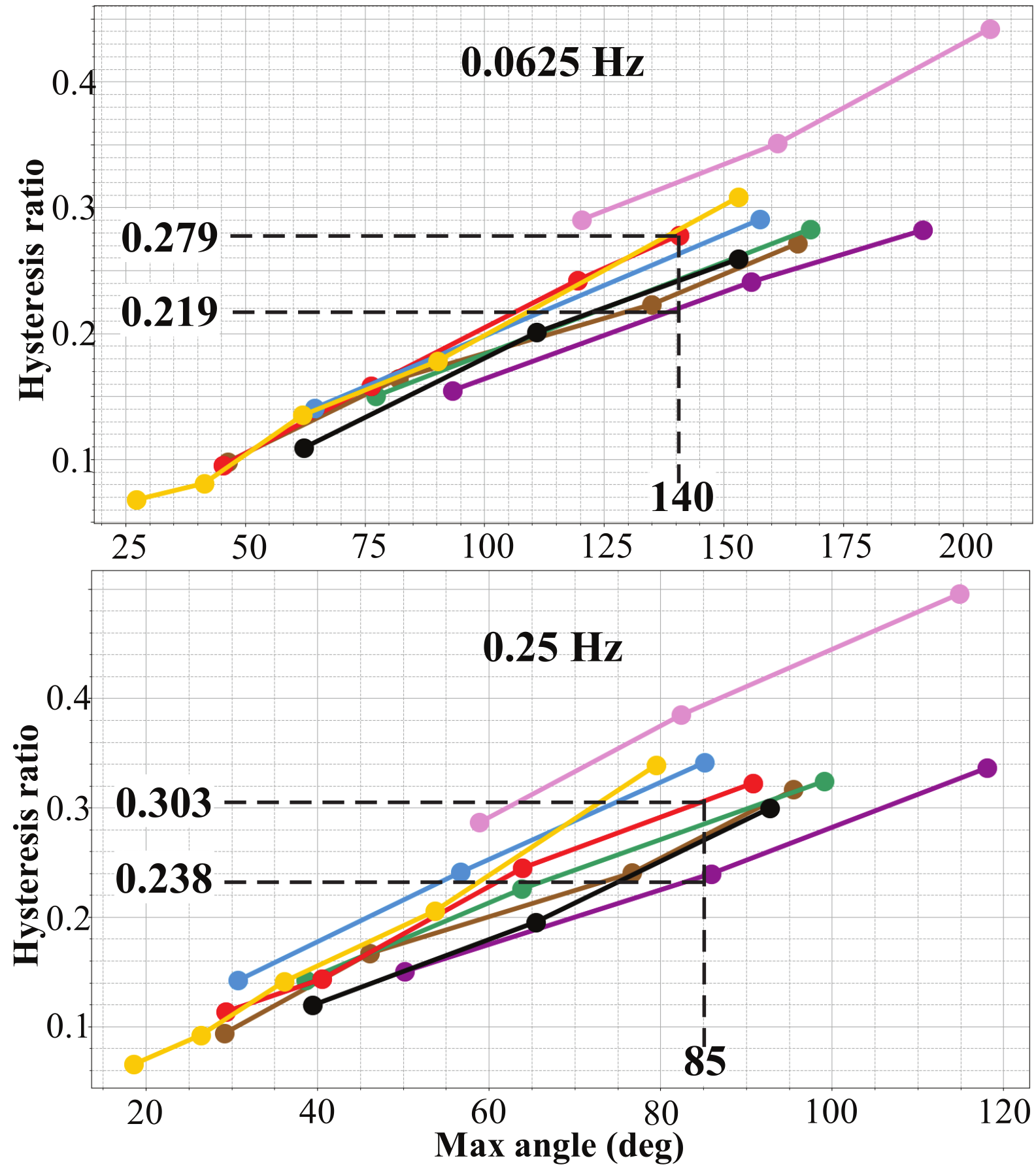}
        \fi    	         
        \label{subfig.dynHystVsAngleBoth}
    }
    \caption{(\protect\subref{subfig.dynHystBar}) The top row shows the maximum bending angle reached by each leaf in Fig.~\ref{fig.dynLeavesAll}. The bottom row shows the hysteresis ratio of each leaf. (\protect\subref{subfig.dynHystVsAngleBoth}) The metrics from \protect\subref{subfig.dynHystBar} are plotted against each other. The dashed black lines depict the interpolated constant angle comparisons of the hysteresis ratio between the original (red) and best performing (purple) designs at an angle of 140$^{\circ}$ for 0.0625Hz and an angle of 85$^{\circ}$ for 0.25Hz.}
    \label{fig.dynHystRatio}
    \vspace{-20pt}
\end{figure}

To make a meaningful comparison between actuators with highly variable dynamic responses, it is necessary to consider the general application, which in this case is robotic control.
For rotary actuators in a robotics application, the goal would be to control the bending angle or some derivative thereof.
One would then want to compare two actuator designs with a constant peak-to-peak bending angle and then compare the cumulative hysteresis normalized by the variable input pressure.
Commensurately one could take the hysteresis ratio to normalize by both input and output.
We cannot explicitly conduct constant peak-to-peak angle comparisons since our system does not have position control.
However, we can implicitly make this comparison by interpolating between angle values reached from our constant pressure experiments, as is done in Fig.~\ref{fig.dynHystRatio}.

Fig.~\ref{subfig.dynHystBar} shows bar plots for metrics calculated from Fig.~\ref{fig.dynLeavesAll}.
Again, low-frequency trials are in the left column and high-frequency in the right.
The colors denote each design, consistent with all previous legends.
The top row of plots shows the maximum angle reached by each leaf in Fig.~\ref{fig.dynLeavesAll}.
The bottom row shows the hysteresis ratio of each leaf.
Note that for the actuator with the lowest dynamic response, \emph{mid-top wall thicker} (yellow), we conduct one last trial at 60 kPa peak-to-peak to get it to a higher bending angle.

These metrics are then used as the x- and y- coordinates of the plots in Fig.~\ref{subfig.dynHystVsAngleBoth}.
The top plot is for the low frequency, and the bottom is for the high frequency.
Each color is for a specific design, consistent with the color legend in previous plots.
So, for example, the purple curve in the top plot shows the metrics for the \emph{mid-width smaller} design for the 0.0625 Hz frequency.
The three data points in this curve, from left to right, correspond to the 40, 45, and 50 kPa trials, respectively.
The points are connected with lines to create a point-wise linear interpolation.
For designs that have 55 and 60 kPa trials, there are fourth and fifth data points as well.
An implicit constant angle comparison can then be conducted for any angle along the x-axis as long as the two actuators being compared reach the given angle.
In the top plot, a vertical dashed line is drawn at an angle of 140$^{\circ}$ to compare the best-performing design (\emph{mid-width smaller}, purple) to the \emph{original} (red) design.
For the \emph{original} (red) design, the vertical line intersects near its last data point at 55 kPa.
This means that at 55 kPa, this design reaches a max angle of 140$^{\circ}$.
For the \emph{mid-width smaller} (purple) design, the vertical line intersects between the first and second data points of 40 and 45 kPa, respectively.
This means that this design reaches 140$^{\circ}$ somewhere between its 40 and 45 kPa pressure trials.
We can implicitly analyze the actuator's behavior at this angle by interpolating the value between the two data points.
In so doing, we can report a hysteresis ratio of 0.279 for the \emph{original} (red) design and 0.219 for the \emph{mid-width smaller} (purple) design.
This equates to a 21.5\% improvement in hysteresis.
Considering the higher 0.25 Hz frequency plot, a comparison is drawn at 85$^{\circ}$, which results in a value of 0.303 for the \emph{original} (red) design and 0.238 for the \emph{mid-width smaller} (purple) design.
This also equates to a 21.5\% improvement.
Furthermore, it can be visualized that for both frequencies, for any given angle, the \emph{mid-width smaller} (purple) design has the lowest hysteresis ratio compared with all other designs.
The improvements at these two exemplary angles of 140$^{\circ}$ and 85$^{\circ}$ are summarized in Table~\ref{table.hystRatio}.

\begin{table}[htb]
	\renewcommand{\arraystretch}{1.3} 
	\caption[]{Improvement of normalized hysteresis of the best-performing design compared to the original design at a given frequency and pk-to-pk angle}\vspace{-5pt}
	\centering
	\begin{tabular}{ l | l | l | l | l}
		\hline
		\textbf{Frequency} & \textbf{Angle} & \textbf{Original} & \textbf{Mid-width} & \textbf{Improvement}\\
		\textbf{} & \textbf{} & \textbf{} & \textbf{Smaller} & \textbf{}\\
		\hline
		\hline
		0.0625 Hz & 140$^{\circ}$ & 0.279 & 0.219 & 21.5\%\\
		\hline
		0.25 Hz & 85$^{\circ}$ & 0.303 & 0.238 & 21.5\%\\
		\hline
	\end{tabular}
	\label{table.hystRatio} 
\end{table}

Not only does the \emph{mid-width smaller} (purple) design improve hysteresis, but it also improves dynamic response for both low and high frequencies.
Table~\ref{table.dynResp} compares the dynamic response of these two actuators, which is taken from the metrics in the top row of Fig.~\ref{subfig.dynHystBar}.

\begin{table}[htb]\vspace{-5pt}
	\renewcommand{\arraystretch}{1.3} 
	\caption[]{Improvement in dynamic response of the best-performing design compared to the original design at a given frequency and pk-to-pk pressure}\vspace{-5pt}
	\centering
	\begin{tabular}{ l | l | l | l | l}
		\hline
		\textbf{Frequency} & \textbf{Pressure} & \textbf{Original} & \textbf{Mid-width} & \textbf{Improvement}\\
		\textbf{} & \textbf{} & \textbf{} & \textbf{Smaller} & \textbf{}\\
		\hline
		\hline
		.0625 Hz & 40 kPa &  45.44$^{\circ}$ &  93.28$^{\circ}$ & 105\%\\
		\hline
		.0625 Hz & 45 kPa &  76.30$^{\circ}$ & 155.83$^{\circ}$ & 104\%\\
		\hline
		.0625 Hz & 50 kPa & 119.48$^{\circ}$ & 191.66$^{\circ}$ &  60\%\\
		\hline
		\hline
		  .25 Hz & 40 kPa &  29.34$^{\circ}$ &  50.19$^{\circ}$ &  71\%\\
		\hline
		  .25 Hz & 45 kPa &  40.57$^{\circ}$ &  85.91$^{\circ}$ & 112\%\\
		\hline
		  .25 Hz & 50 kPa &  63.92$^{\circ}$ & 118.06$^{\circ}$ &  85\%\\
		\hline
		\hline
	\end{tabular}
	\label{table.dynResp} 
\end{table}

\vspace{-10pt}
\section{Discussion}\label{sec.discussion}
\begin{figure}[htb]
    \if\whichFig0
        \includegraphics[width=\linewidth]{figures/aiEps/staticStressMiddleBar.eps}
    \fi
    \if\whichFig1
        \includegraphics[width=\linewidth]{figures/aiTif/staticStressMiddleBar.tif}
    \fi
    \if\whichFig2
        \includegraphics[width=\linewidth]{figures/aiPdf/staticStressMiddleBar.pdf}
    \fi    	
    \if\whichFig3
    	\includegraphics[width=\linewidth]{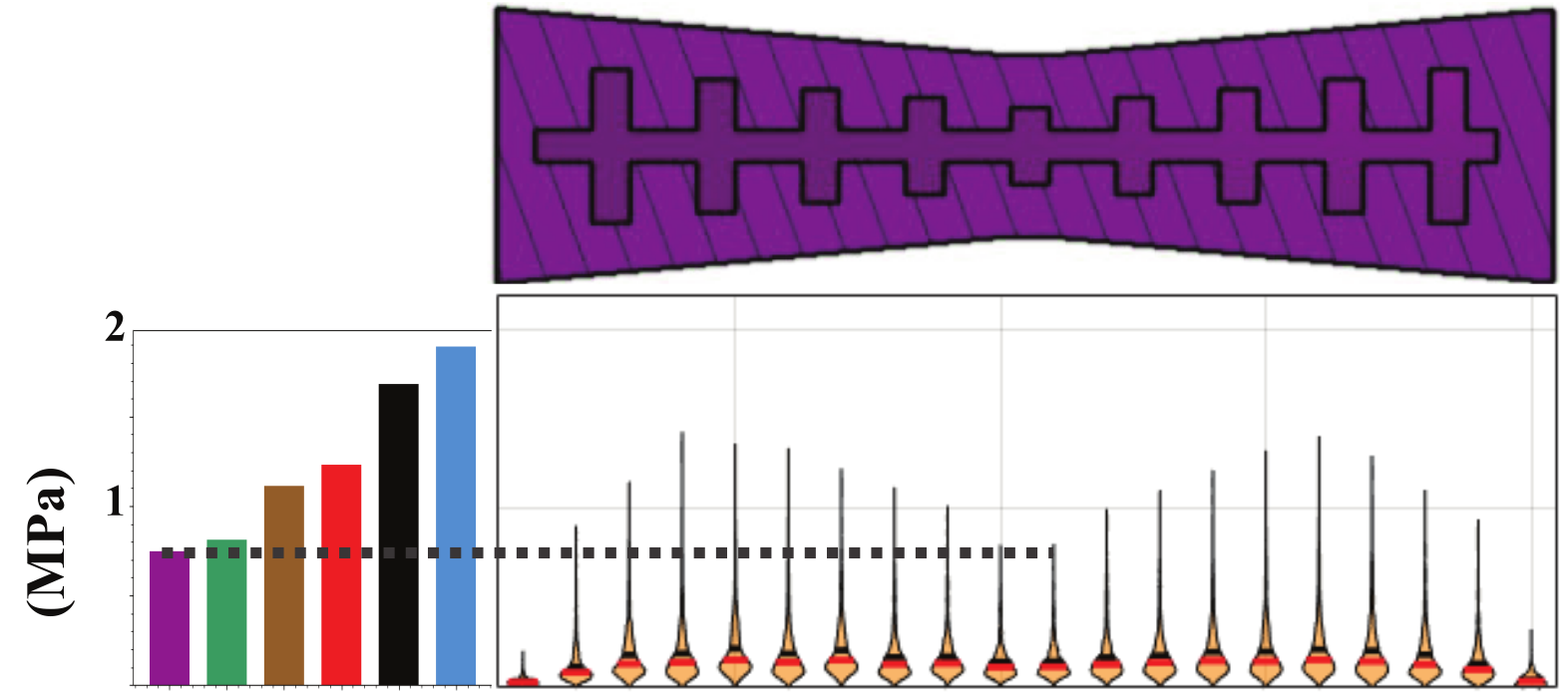}
    \fi    	         
    \caption{\emph{(Left)} Bar chart of the max von Mises stress of the center cross-sectional slice of the five actuators from Fig.~\ref{fig.staticStressDistributionsAll}. \emph{Right} Corresponding distribution of the best performing (purple) actuator, to explain in more detail the purple bar on the left.}
    \label{fig.staticStressMiddleBar}
    \vspace{-10pt}
\end{figure}

\if\whichHyp0
In this section, we discuss the correlation between stress distribution and dynamic performance.
\fi
\if\whichHyp1
In this section, we discuss the correlation between stress distribution and dynamic performance.
\fi
The \emph{mid-width smaller} (purple) design is the best-performing actuator with regards to dynamic metrics (normalized hysteresis and dynamic response) for both high and low frequencies.
We observe from the static analysis in Fig.~\ref{fig.staticStressDistributionsAll} that the \emph{mid-width smaller} (purple) design also has the lowest stress in the center.
Fig.~\ref{fig.staticStressMiddleBar} shows a bar chart (\emph{left}) which plots the maximum value of the stress distribution in the center cross-sectional slice for each design.
Fig.~\ref{fig.staticStressMiddleBar} \emph{right} shows the corresponding cross-sectional stress distributions for the best-performing design.
Note that it is not the \emph{overall} maximum stress that is correlated with the dynamics, but rather the maximum stress of the \emph{center region}.
\if\whichHyp0
This suggests that minimizing stress at the actuator's center, which in turn minimizes the stored potential energy at the center, can lessen the energy dissipation, as observed with the lower hysteretic behavior.
\fi
\if\whichHyp1
This supports our hypothesis that decentralizing the stress distribution, which in turn decentralizes stored potential energy, can alter the energy dissipation, as observed with the lower hysteretic behavior.
In particular, our experiments show that decentralizing the stress distribution alters the hysteresis by \emph{lowering} it.
\fi

Reflecting on the FEM analysis in Section~\ref{subsec.femResults}, we notice how the \emph{mid-width smaller} (purple) design stands out as unique.
Considering the bending angle, it can be observed in Fig.~\ref{fig.staticFemAngleVsPressure} that for the air chamber width- and height-varying designs, a smaller volume in the center produces a higher bending angle (purple and brown), while a larger volume in the center produces a lower bending angle (black and orange).
This trend is not observed for the length-varying designs (green and blue), which remain close to the original (red).
Considering the stress distribution, it can be observed in Fig.~\ref{fig.staticStressDistributionsAll} that for the air chamber width- and length-varying designs, a smaller volume in the center produces less stress in the center (purple and green), while a larger volume in the center produces more stress in the center (black and blue).
This trend is not observed for the height-varying design (brown).
So, the \emph{mid-width smaller} (purple) is the only design that optimizes \emph{both} the bending angle and stress distribution.
This suggests that the union of the optimization of these two static metrics (bending angle and stress distribution) correlates with the optimization of the two dynamic metrics (hysteretic and dynamic response).
It is also interesting to consider that the width parameter is on the transverse (x-) axis, which is the axis normal to the bending plane.
These are benchmarks that future researchers can use for further investigation and design optimization.

\section{Conclusion}\label{sec.conclusion}


This paper explores the impact of the internal geometrical structure of soft pneu-nets on the actuator's dynamic response and hysteresis.
The study demonstrates that by particular manipulation of the stress distribution within soft robots, it is feasible to boost the dynamic response while reducing hysteresis.
For this, the study employs Finite Element Modeling (FEM) for up to 500\% strain of the actuator (with 95\% accuracy for predicting the bending angle) and experimental validation through markerless motion tracking of the soft robot.
The paper evaluates the effectiveness of "mechanically programming" stress distribution and distributed energy storage inside soft robots to maximize dynamic performance, thereby offering direct benefits for control. 
As such, we demonstrate that the \emph{mid-width smaller} (purple) design improves hysteretic behavior by 21.5\% for both high and low frequencies and improves dynamic response by 60\% to 112\% for various frequencies and peak-to-peak pressures.

\section*{Acknowledgments}
The authors would like to acknowledge the help of Mr. Aadit Patel and Mr. Naigam Bhatt for their help with the motion capture part of the experimental setup. This material is based upon work supported in part by the US National Science Foundation under grants no \#2121391 and \#2208189. The work is also partially supported by NYUAD CAIR award \# CG010. 



\bibliographystyle{./IEEEtran}
\bibliography{./IEEEabrv,./references}


\section{Biography Section}
 

\vspace{-30pt}
\begin{IEEEbiography}%
[{%
    \if\whichFig0
        \includegraphics[width=1in,height=1.50in,clip,keepaspectratio]%
            {figures/aiEps/headshot-orig-libby.eps}%
    \fi
    \if\whichFig1
        \includegraphics[width=1in,height=1.50in,clip,keepaspectratio]%
            {figures/aiTif/headshot-orig-libby.tif}%
    \fi
    \if\whichFig2
        \includegraphics[width=1in,height=1.50in,clip,keepaspectratio]%
            {figures/aiPdf/headshot-orig-libby.pdf}%
    \fi    	
    \if\whichFig3
    	\includegraphics[width=1in,height=1.50in,clip,keepaspectratio]%
	    {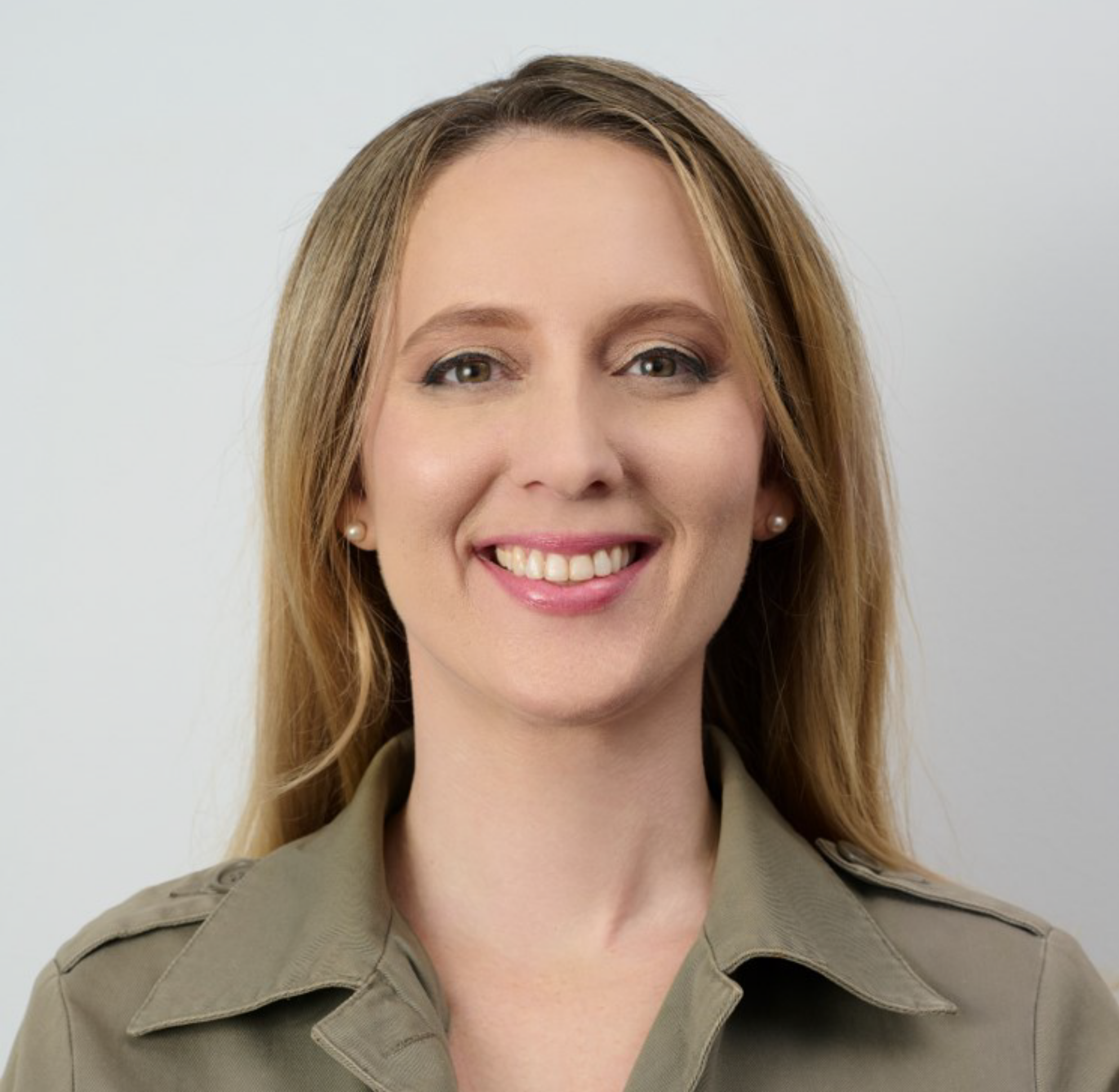}%
    \fi    	         
}]%
{Jacqueline Libby}
(Member, IEEE) received the Ph.D. degree in Robotics from Carnegie Mellon University, Pittsburgh, PA, USA in 2019. She is currently a Postdoctoral Associate at New York University (NYU) Tandon School of Engineering, New York, NY, USA, where she has been awarded a two-year fellowship through the NYU Center for Urban Science and Progress (CUSP). Her research interests include rehabilitation robotics, soft robotics, machine learning, human-machine interfaces, and human-centric mechanical design.
\end{IEEEbiography}

\vspace{-30pt}
\begin{IEEEbiography}%
[{%
    \if\whichFig0
        \includegraphics[width=1in,height=1.50in,clip,keepaspectratio]%
            {figures/aiEps/headshot-orig-somwanshi.eps}%
    \fi
    \if\whichFig1
        \includegraphics[width=1in,height=1.50in,clip,keepaspectratio]%
            {figures/aiTif/headshot-orig-somwanshi.tif}%
    \fi
    \if\whichFig2
        \includegraphics[width=1in,height=1.50in,clip,keepaspectratio]%
            {figures/aiPdf/headshot-orig-somwanshi.pdf}%
    \fi    	
    \if\whichFig3
    	\includegraphics[width=1in,height=1.50in,clip,keepaspectratio]%
	    {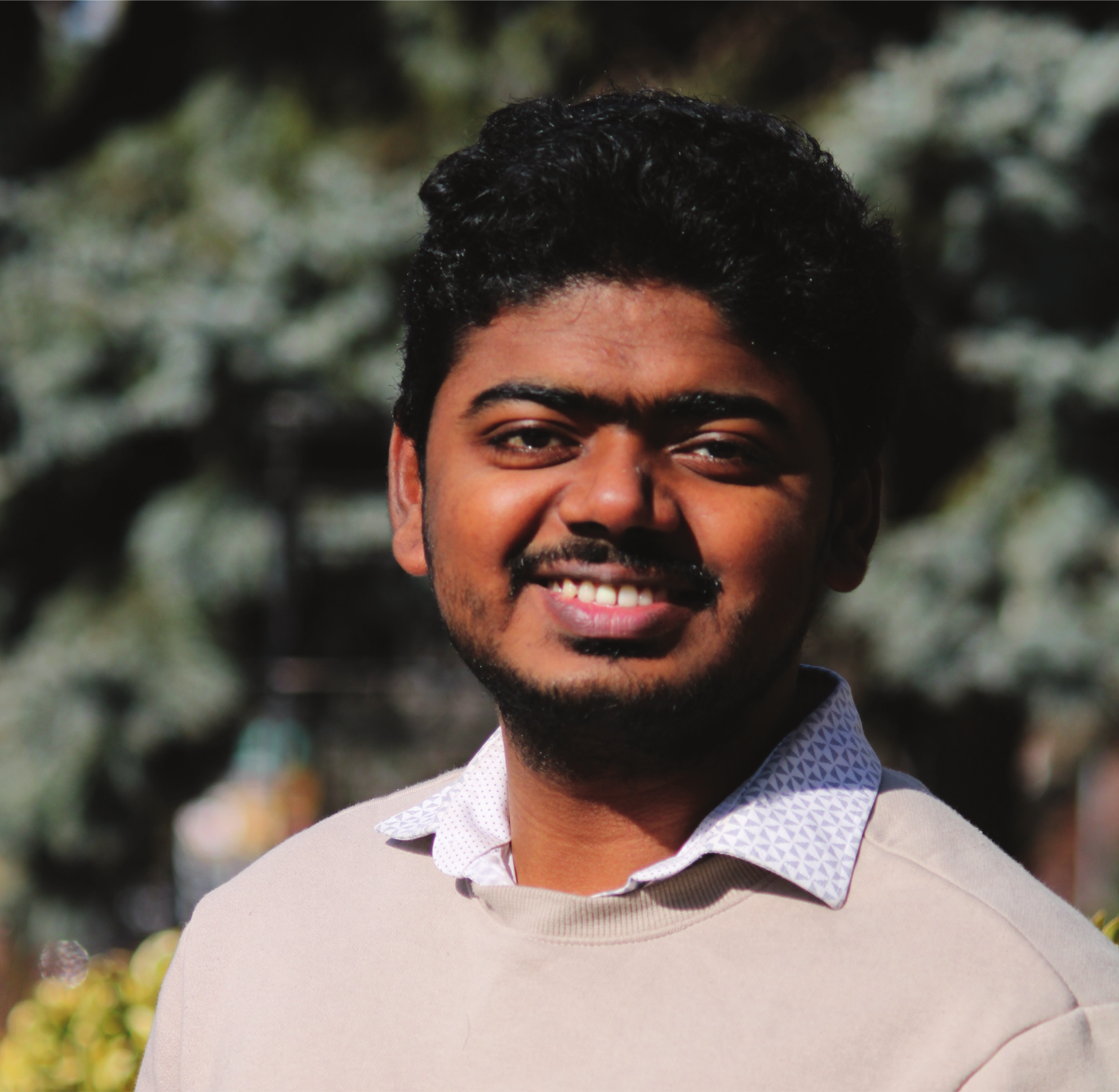}%
    \fi    	         
}]%
{Aniket A. Somwanshi}
received his M.S. in Mechatronics and Robotics from NYU Tandon School of Engineering in 2022, and a B.E. in Mechanical Engineering from Savitribai Phule Pune University, India. His masters research at NYU was focused on the Design of Soft Actuators. He currently works as a Robotics Firmware Engineer at Azure Medical Innovations. His research interests include exoskeletons, simulation and controls, and surgical robotics.
\end{IEEEbiography}

\begin{IEEEbiography}%
[{%
    \if\whichFig0
        \includegraphics[width=1in,height=1.50in,clip,keepaspectratio]%
            {figures/aiEps/headshot-orig-stancati.eps}%
    \fi
    \if\whichFig1
        \includegraphics[width=1in,height=1.50in,clip,keepaspectratio]%
            {figures/aiTif/headshot-orig-stancati.tif}%
    \fi
    \if\whichFig2
        \includegraphics[width=1in,height=1.50in,clip,keepaspectratio]%
            {figures/aiPdf/headshot-orig-stancati.pdf}%
    \fi    	
    \if\whichFig3
    	\includegraphics[width=1in,height=1.50in,clip,keepaspectratio]%
	    {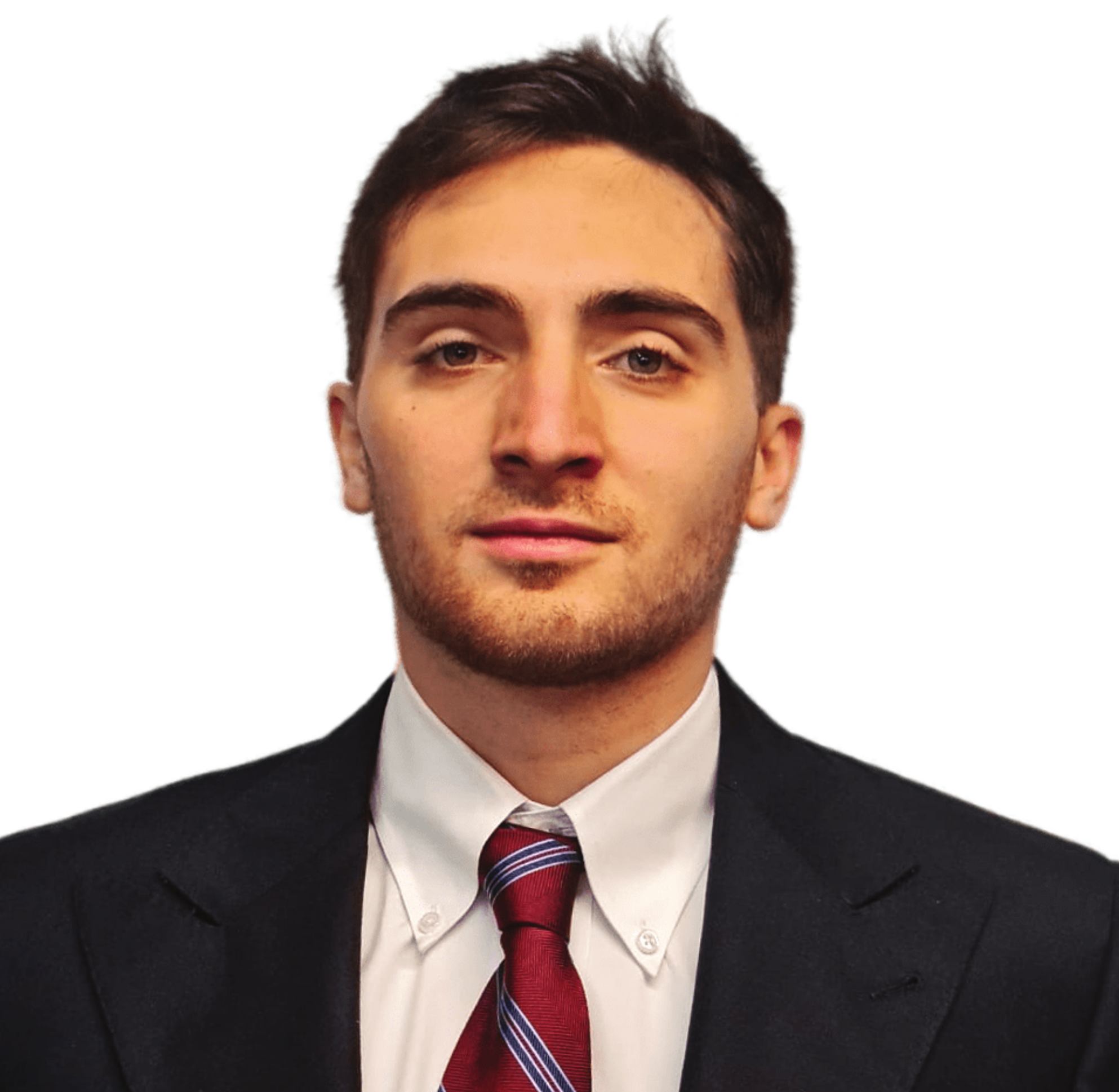}%
    \fi    	         
}]%
{Federico Stancati}
completed his B.Sc. in Mechanical Engineering at La Sapienza University, Rome, Italy. He received a joint M.S. between NYU and La Sapienza in Mechanical Engineering in 2023. His masters research focused on the development of reliable FEM methods to simulate the behavior of hyperelastic actuators. Applications include healthcare, rehabilitation, and exoskeletons.
\end{IEEEbiography}

\vspace{-40pt}
\begin{IEEEbiography}%
[{
    \if\whichFig0
        \includegraphics[width=1in,height=1.50in,clip,keepaspectratio]%
            {figures/aiEps/headshot-orig-tyagi.eps}%
    \fi
    \if\whichFig1
        \includegraphics[width=1in,height=1.50in,clip,keepaspectratio]%
            {figures/aiTif/headshot-orig-tyagi.tif}%
    \fi
    \if\whichFig2
        \includegraphics[width=1in,height=1.50in,clip,keepaspectratio]%
            {figures/aiPdf/headshot-orig-tyagi.pdf}%
    \fi    	
    \if\whichFig3
    	\includegraphics[width=1in,height=1.50in,clip,keepaspectratio]%
	    {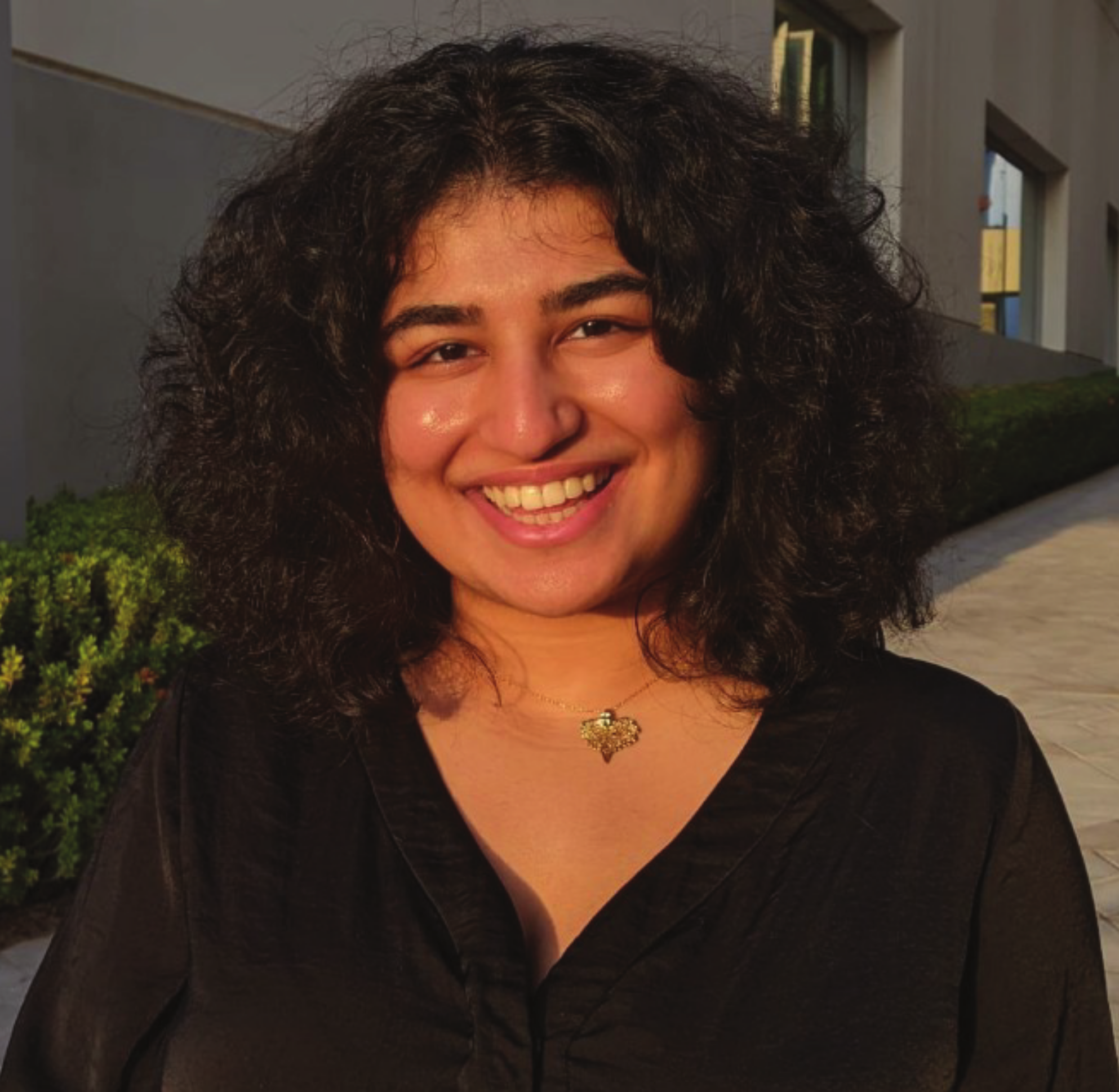}%
    \fi    	         
}]%
{Gayatri Tyagi}
is an undergraduate student at NYU Tandon School of Engineering, studying computer science, with minors in robotics, mathematics, and mechanical engineering. She is a member of the Global Leaders and Scholars in STEM honors program at NYU Tandon and conducts research in the MERIIT lab. She is passionate about developing technologies that solve existing problems and help improve people's lives. 
\end{IEEEbiography}

\vspace{-40pt}
\begin{IEEEbiography}%
[{%
    \if\whichFig0
        \includegraphics[width=1in,height=1.50in,clip,keepaspectratio]%
            {figures/aiEps/headshot-orig-mehrdad.eps}%
    \fi
    \if\whichFig1
        \includegraphics[width=1in,height=1.50in,clip,keepaspectratio]%
            {figures/aiTif/headshot-orig-mehrdad.tif}%
    \fi
    \if\whichFig2
        \includegraphics[width=1in,height=1.50in,clip,keepaspectratio]%
            {figures/aiPdf/headshot-orig-mehrdad.pdf}%
    \fi    	
    \if\whichFig3
    	\includegraphics[width=1in,height=1.50in,clip,keepaspectratio]%
	    {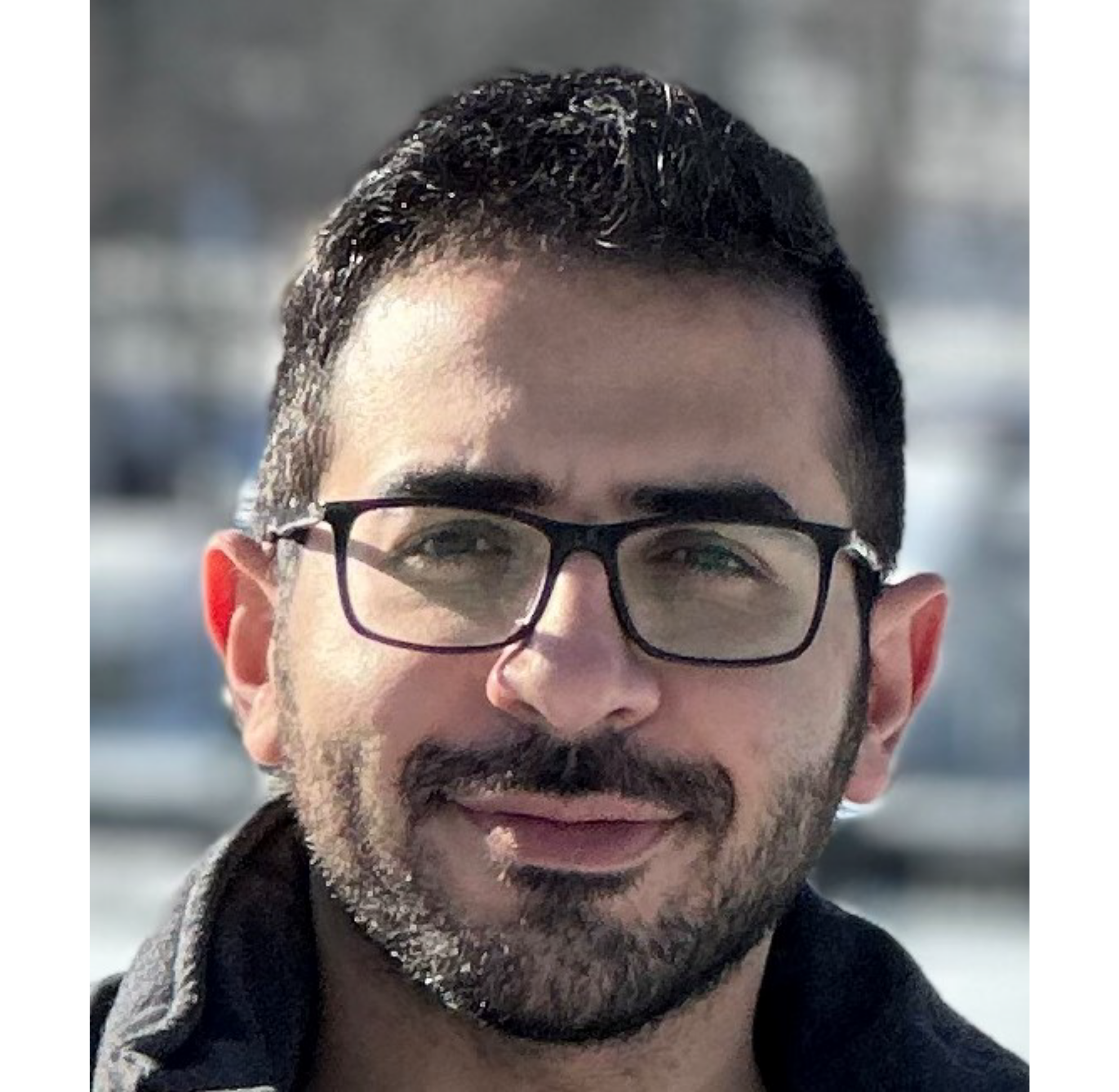}%
    \fi    	         
}]%
{Sarmad Mehrdad}
received his B.Sc. in Mechanical Engineering from K. N. Toosi University, Tehran, Iran, and his M.S. in Mechanical Engineering from Clarkson University, Potsdam, NY, USA. He is a Ph.D. candidate in Electrical Engineering at NYU Tandon under the supervision of Prof. S. F. Atashzar. His research includes learning from demonstration on robotic platforms, wearables for biomarker analysis, prognostic deep learning, and soft robotics. He has received fellowships from NYU Tandon and the NYU K-12 STEM Educational Program.
\end{IEEEbiography}

\vspace{-30pt}
\begin{IEEEbiography}%
[{%
    \if\whichFig0
        \includegraphics[width=1in,height=1.50in,clip,keepaspectratio]%
            {figures/aiEps/headshot-orig-rizzo.eps}%
    \fi
    \if\whichFig1
        \includegraphics[width=1in,height=1.50in,clip,keepaspectratio]%
            {figures/aiTif/headshot-orig-rizzo.tif}%
    \fi
    \if\whichFig2
        \includegraphics[width=1in,height=1.50in,clip,keepaspectratio]%
            {figures/aiPdf/headshot-orig-rizzo.pdf}%
    \fi    	
    \if\whichFig3
    	\includegraphics[width=1in,height=1.50in,clip,keepaspectratio]%
	    {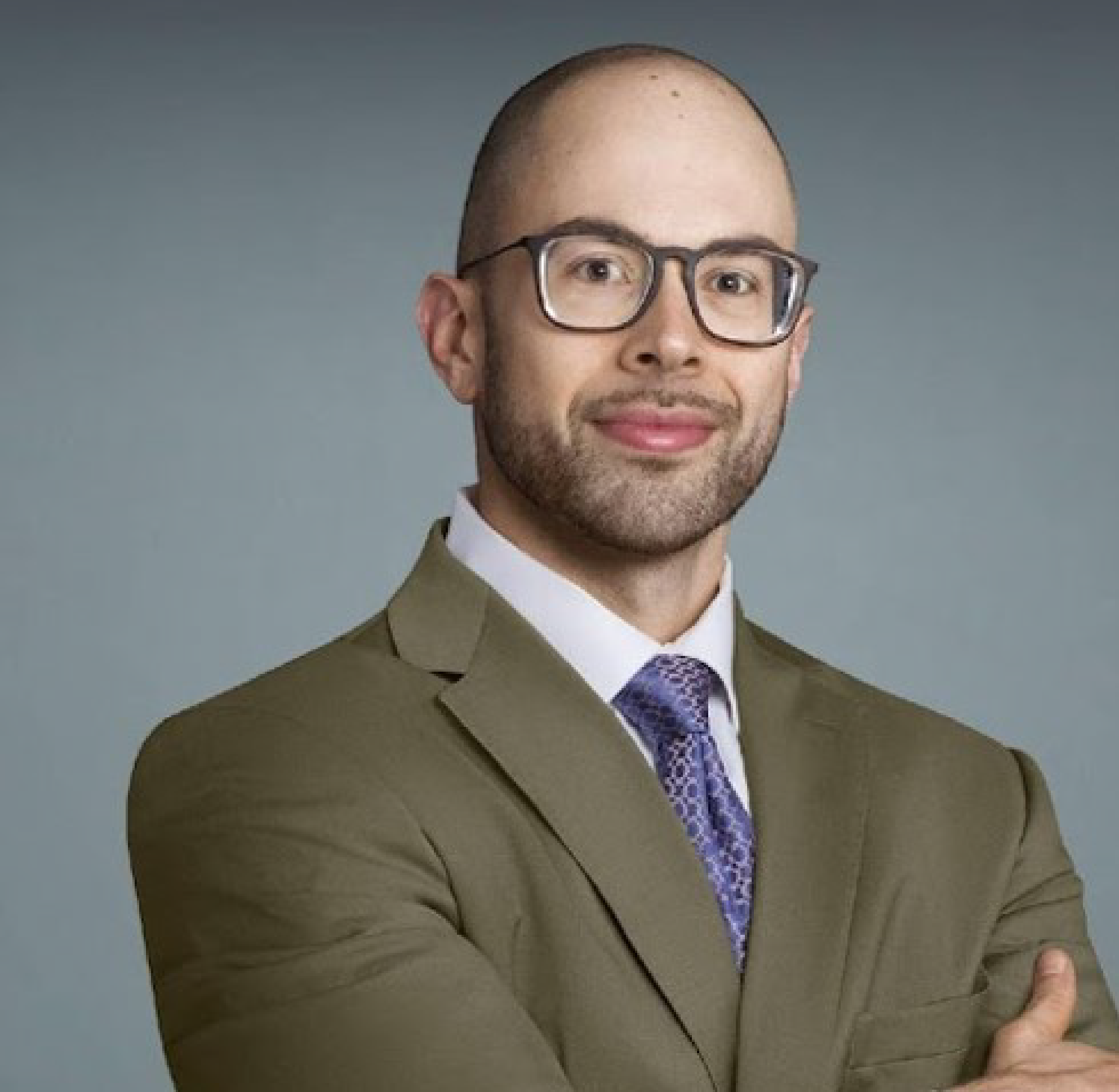}%
    \fi    	         
}]%
{John-Ross (JR) Rizzo}
, M.D., M.S.C.I., is a physician-scientist at NYU Langone Medical Center. He is an Associate Professor, serving as Health System Director of Disability Inclusion at NYU Langone Medical Center and Vice chair of Innovation and Equity for the Department of Physical medicine and rehabilitation at Rusk Institute of Rehabilitation Medicine, with cross-appointments in the Department of Neurology and the Departments of Biomedical \& Mechanical and Aerospace Engineering New York University Tandon School of Engineering. He is also the Associate Director of Healthcare for the NYU WIRELESS Laboratory in the Department of Electrical and Computer Engineering at New York University Tandon School of Engineering. He leads the Visuomotor Integration Laboratory (VMIL), where his team focuses on eye-hand coordination, as it relates to acquired brain injury, and the REACTIV Laboratory (Rehabilitation Engineering Alliance and Center Transforming Low Vision), where his team focuses on advanced wearables for the sensory deprived and benefits from his own personal experiences with vision loss.
\end{IEEEbiography}

\vspace{-30pt}
\begin{IEEEbiography}%
[{%
    \if\whichFig0
        \includegraphics[width=1in,height=1.50in,clip,keepaspectratio]%
            {figures/aiEps/headshot-orig-atashzar.eps}%
    \fi
    \if\whichFig1
        \includegraphics[width=1in,height=1.50in,clip,keepaspectratio]%
            {figures/aiTif/headshot-orig-atashzar.tif}%
    \fi
    \if\whichFig2
        \includegraphics[width=1in,height=1.50in,clip,keepaspectratio]%
            {figures/aiPdf/headshot-orig-atashzar.pdf}%
    \fi    	
    \if\whichFig3
    	\includegraphics[width=1in,height=1.50in,clip,keepaspectratio]%
	    {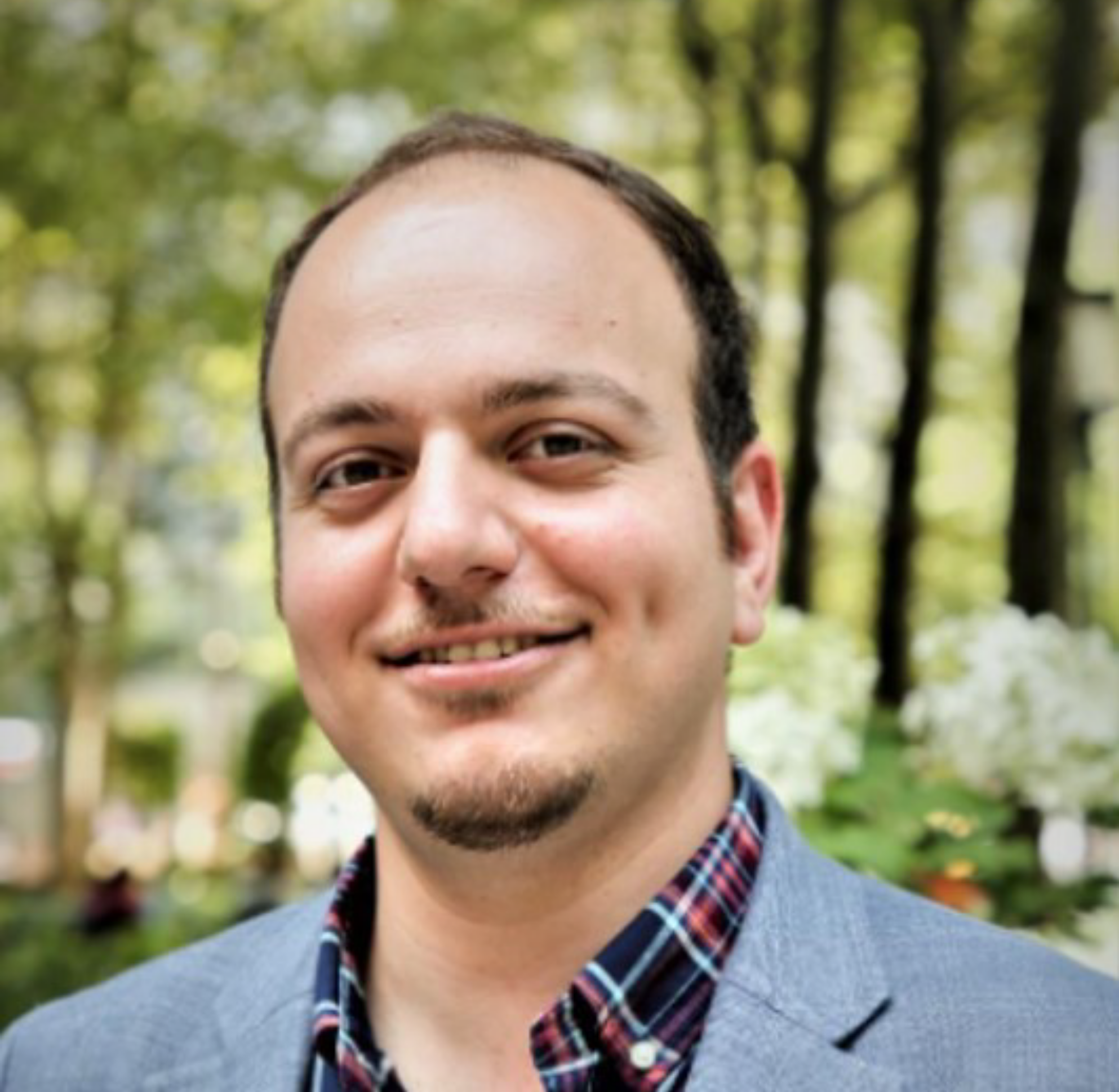}%
    \fi    	         
}]%
{S. Farokh Atashzar}
(Senior Member, IEEE) received the Ph.D. degree in electrical and computer engineering from the University of Western Ontario, Canada, in 2017.
He is currently an Assistant Professor with New York University (NYU), New York, NY, USA, jointly appointed with the Department of Electrical and Computer Engineering and Mechanical and Aerospace Engineering. He is also affiliated with NYU WIRELESS, New York, and NYU Center for Urban Science and Progress, New York, and Leads the Medical Robotics and Interactive Intelligent Technologies (MERIIT) Lab, NYU, and the activities of the Lab are funded by the U.S. National Science Foundation. Prior to joining NYU, he was a Postdoctoral Scientist at Imperial College London, U.K. His research interests include human-machine interfaces, haptics, human-centered robotics, biosignal processing, deep learning, and nonlinear control. He was the recipient of several awards, including the 2021 Outstanding Associate Editor of IEEE RAL. He is an Associate Editor for IEEE Transactions on Robotics, IEEE Transactions on Haptics and IEEE RAL.
\end{IEEEbiography}



\end{document}